\documentclass[journal]{IEEEtran}
\usepackage{amsthm}
\usepackage{url}
\usepackage{cite}
\usepackage{amsmath,amssymb,amsfonts}
\usepackage{algorithmic}
\usepackage{graphicx}
\graphicspath{{./}}
\usepackage[linewidth=1pt]{mdframed}
\usepackage{mathtools}
\usepackage{dsfont}
\usepackage[dvipsnames]{xcolor}
\usepackage{siunitx}
\usepackage{booktabs}
\usepackage{hyperref}
\usepackage{cleveref}
\usepackage{tikz}
\usetikzlibrary{shapes,arrows,fit,positioning}
\usepackage{balance}
\usepackage[most]{tcolorbox}
\Crefname{figure}{Fig.}{Figs.}
\tcbset{on line,
        boxsep=4pt,
        left=6pt, right=6pt, top=6pt, bottom=6pt}

\newtheorem{theorem}{Theorem}[section]

\newtheorem{proposition}[theorem]{\revdel{Theorem}\revadd{Proposition}}
\newtheorem{remark}{Remark}[section]

\newif\ifshowrevision
\showrevisionfalse
\DeclareRobustCommand{\revadd}[1]{\ifshowrevision{\color{blue}\relax#1}\else#1\fi}
\DeclareRobustCommand{\revdel}[1]{\ifshowrevision{\color{red}\relax#1}\ifmmode\;\else\space\fi\fi}
\makeatletter
\def\@IEEEsecHeadingPunct{.}

\def\subsection{\@startsection{subsection}{2}{\parindent}{0ex plus 0.1ex minus 0.1ex}{0ex}{\normalfont\normalsize\bfseries}}
\makeatother

\usepackage{etoolbox}

\makeatletter

\def\@IEEEsectpunct{. }   % <-- add this
\def\subsection{\@startsection{subsection}{2}{\parindent}{0ex plus 0.1ex minus 0.1ex}{0ex}{\normalfont\normalsize\bfseries}}
\makeatother

\title{\LARGE \bf
BRACE: Belief-guided Residual Adaptive Control\\
for Safe Navigation under Unknown Disturbances
}

\author{Jaeyoun Choi\textsuperscript{*}, Niklas Schmid\textsuperscript{*}, Oswin So, Augustinos Saravanos, and Chuchu Fan% <-this % stops a space
\thanks{\textsuperscript{*}Jaeyoun Choi and Niklas Schmid contributed equally to this work.}%
\thanks{Jaeyoun Choi, Oswin So, Augustinos Saravanos, and Chuchu Fan are with the
Reliable Autonomous Systems Lab, Massachusetts Institute of Technology,
Cambridge, MA 02139, USA
(e-mail: \{cjy0051, oswinso, asaravan, chuchu\}@mit.edu).}%
\thanks{Niklas Schmid is with the Automatic Control Laboratory (IfA),
ETH Zürich, 8092 Zürich, Switzerland
(e-mail: nikschmid@ethz.ch).}%
}
\begin{document}

\maketitle
% \thispagestyle{empty}
% \pagestyle{empty}

%%%%%%%%%%%%%%%%%%%%%%%%%%%%%%%%%%%%%%%%%%%%%%%%%%%%%%%%%%%%%%%%%%%%%%%%%%%%%%%%
%  First sentence. unknown disturbance -> not clear. spatially varying can be part. Too general. SHortcomings. Problem. To address This is hard problem. 
% Safe autonomous navigation is challenging under spatially varying disturbance in cluttered environments. This problem requires adapting not only how a trajectory is executed, but also which trajectory should be planned. Existing planning-layer approaches typically require known or bounded disturbance models.
\begin{abstract}
\revadd{Disturbances encountered during autonomous navigation generally vary across space, often discontinuously, as obstacles redirect wind and changing terrain alters traction.}\revdel{Safe autonomous navigation under spatially varying disturbances in
cluttered environments is challenging.} Tracking controllers typically
assume sufficient control authority to mitigate disturbances and maintain safety; the regime
where this assumption fails remains underexplored. \revdel{In such cases, the
problem requires adapting not only how a trajectory is executed, but
also how this trajectory is originally planned.}\revdel{However, existing planning-layer approaches either rely on conservative assumptions for tractable optimization, or assume disturbances with known structure.} We present Belief-guided Residual Adaptive
Control (BRACE), a hierarchical reinforcement learning (RL) framework for
reach-avoid navigation under \revadd{unknown, bounded,} spatially varying disturbances. BRACE
infers a disturbance belief from model prediction errors
and uses that to condition a high-level policy that adapts the reference trajectory.\revdel{To train the nominal and residual policies
jointly, we introduce a likelihood-ratio advantage reweighting, which assigns
credit to each policy based on its contribution to the sampled action. We train the policy with a control barrier function (CBF) based reinforcement learning approach for reach-avoid behavior.} \revadd{The policy combines a nominal and a belief-guided residual term, trained jointly with a likelihood-ratio advantage reweighting heuristic.} We evaluate BRACE in three simulated quadcopter environments with abruptly switching wind fields, and validate it on Crazyflie hardware with fan-generated disturbances. Across these settings, BRACE improves reach-avoid performance over \revdel{tracking-layer disturbance compensation and
RL baselines without disturbance belief, supporting belief-guided residual planning
as a practical mechanism for safe navigation under unknown, spatially
varying disturbances.}\revadd{model predictive control and RL baselines.}
\end{abstract}

%%%%%%%%%%%%%%%%%%%%%%%%%%%%%%%%%%%%%%%%%%%%%%%%%%%%%%%%%%%%%%%%%%%%%%%%%%%%%%%%
\section{Introduction}
\label{sec:intro}

Autonomous robots operating in the real world must act safely under
unknown, spatially varying disturbances,
such as wind fields for aerial vehicles \cite{schmid2022real}. \revdel{Such disturbances are difficult to model and typically cannot be anticipated, requiring both real-time estimation and real-time adaptation. This makes safe navigation especially challenging in cluttered environments, where the safe region might be narrow.}\revadd{Such disturbances can cause deviations from planned trajectories, making safe navigation particularly challenging in cluttered environments, where feasible corridors may be narrow.}
The dominant paradigm for handling disturbances operates at the
trajectory-tracking layer \revadd{through robust and adaptive
control methods~\cite{hovakimyan2010l1, wu2024l1quad,
huang2023datt, ostafew2016robust}.}\revdel{, which assumes sufficient control authority to
reject the disturbance while following a prescribed
reference~\cite{hovakimyan2010l1, hewing2020learning, abdul2025nonlinear}.} \revdel{These tracking-layer methods are fundamentally reactive: they can improve tracking of a given reference, but cannot change a reference that is infeasible under the disturbance and actuation limits.}\revadd{These tracking-layer methods
can improve tracking of a given reference, but cannot
necessarily resolve reference infeasibility under disturbances
and actuation limits. This limitation motivates adapting reference
trajectories at the planning level.

Hierarchical control frameworks provide a natural architecture
for this purpose by separating high-level reference planning
from low-level trajectory tracking, allowing references to be adapted while retaining a dedicated tracking controller. One approach bounds the
tracking errors of low-level controllers and plans references
with corresponding safety margins around obstacles~\cite{fan2020fast}. This allows planning references that
remain safe under bounded tracking errors. However, predetermined safety margins based on worst-case tracking errors may become overly conservative in cluttered environments, particularly when disturbances exhibit spatial and temporal correlations that can be exploited through online estimation.}

\revdel{This limitation highlights the need to
accommodate such uncertainties at the planning level, i.e., using disturbance estimates to modify the reference trajectories themselves. Several recent works
adapt the planned motion through optimization-based
replanning}\revadd{Building upon this motivation, recent approaches suggest
adapting references to online disturbance estimates through
optimization-based replanning}~\cite{wu2021external, zhang2024change, wang2025seal,
lapandic2024metampc}\revadd{.}\revdel{, but} \revadd{However, }real-time re-optimization typically relies on
simplified dynamics, bounded uncertainty sets, or reachable-set
approximations~\cite{herbert2017fastrack}, \revdel{which grow conservative when
obstacles leave narrow feasible corridors}\revadd{which can introduce computational or modeling limitations, particularly in cluttered environments}. Learning-based planners instead amortize planning across many configurations, yet existing
disturbance-adaptation methods assume simplified wind models tied to known geometry, or modify waypoints for a fixed, pre-specified
disturbance~\cite{wu2024multi, ahmed2023rl}. Existing methods do not directly address online reference adaptation to unknown, spatially varying disturbances in cluttered environments.

\begin{figure}[t]
    \centering
    \includegraphics[width=0.9\linewidth]{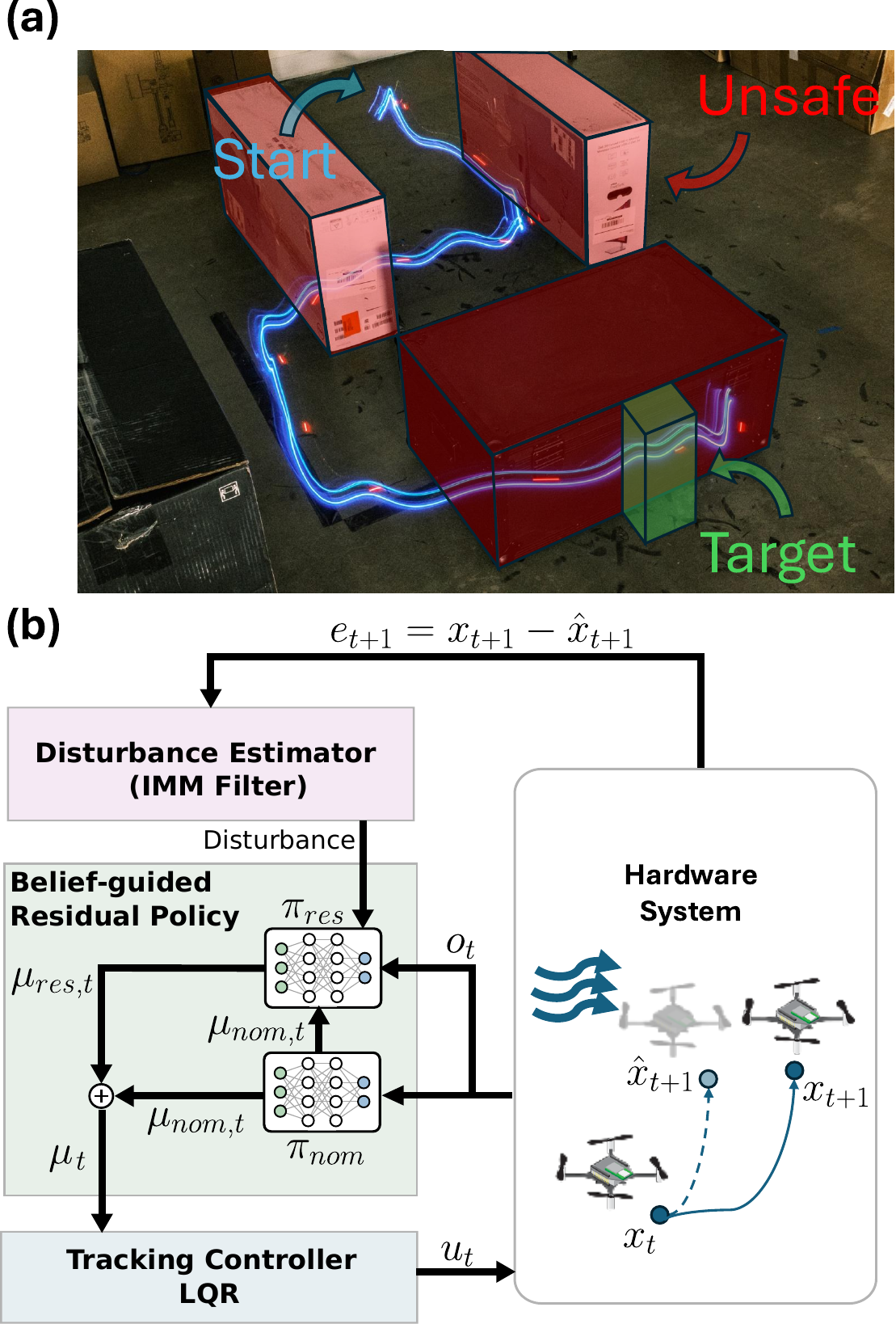}
    \vspace{-0.2cm}
    \caption{
    BRACE reach-avoid experiment and pipeline.
    (a) Representative Crazyflie trajectory through the cluttered workspace.
    (b) Tracking error updates the IMM belief, the belief-guided residual
    policy generates a reference, and the low-level LQR tracks it on hardware.
    }
    \vspace{-0.5cm}
    \label{fig:overview}
\end{figure}

\revdel{  To address these limitations, we introduce \textbf{BRACE}
  (Belief-guided Residual Adaptive Control), a hierarchical RL
  framework for safe trajectory planning
  % \revadd{reference adaptation}
  under unknown
  % \revadd{spatially varying}
  disturbances
  (Fig.~\ref{fig:overview}b). BRACE infers an online disturbance
  belief from model prediction errors using
  an
  % \revadd{a two-mode Gaussian}
  Interacting Multiple Model (IMM) estimator,
  and employs this belief to adapt the high-level reference trajectory.

% MODIFIED: removed duplicate "The main contributions are:" line
The main contributions of this paper are:
\begin{itemize}
    \item \emph{Belief-guided Residual Adaptive Control.}
    We propose BRACE, an approach that decomposes high-level
    motion into a nominal action and a belief-guided residual,
    separating nominal planning from the disturbance-aware adaptation of the reference trajectory.
    % \revadd{jointly learning a nominal branch and a belief-conditioned residual for reference adaptation.}
    We show that BRACE outperforms both tracking-layer compensation 
    % \revadd{and planning-layer MPC}
    given the same belief
    % \revadd{, as well as}
    and RL planning without belief.

    \item \emph{Credit-aware joint training.}
    We introduce likelihood-ratio advantage reweighting
    % \revadd{, a training heuristic,}
    for joint proximal policy optimization (PPO)
    training, assigning separate update weights to the nominal and
    residual policies according to how much each policy explains the
    sampled action. This enables stable
joint training of both policies and outperforms conventional
residual training with a frozen nominal policy.
% \revadd{This supports joint
% training of both policies; BRACE achieves higher mean success than
% Frozen-nominal in three of the four synthetic conditions.}

    \item \emph{Evaluation under spatially varying disturbances.}
    We evaluate BRACE in cluttered reach-avoid tasks with abruptly changing wind fields, and validate the
    learned policy on hardware experiments.
\end{itemize}
}

\revadd{To address this gap, we 
\begin{itemize}
    \item
    introduce Belief-guided Residual Adaptive Control (BRACE), a hierarchical reinforcement learning (RL) framework for reach-avoid navigation under unknown \revadd{spatially varying} disturbances (Fig.~\ref{fig:overview}b).
    \item propose an efficient training heuristic that facilitates joint learning of a nominal policy and a residual policy that adjusts the reference based on online disturbance estimates. 
    \item evaluate our framework in simulation as well as on real-world Crazyflies, and demonstrate that our method outperforms model predictive control and RL baselines.
\end{itemize}
}

%%%%%%%%%%%%%%%%%%%%%%%%%%%%%%%%%%%%%%%%%%%%%%%%%%%%%%%%%%%%%%%%%%%%%%%%%%%%%%%%
\section{Related Work}
\label{sec:related}

\subsection{Robust Tracking Control}
Disturbances are most commonly handled at the low-level control layer,
which improves how a prescribed reference is executed. Adaptive
controllers~\cite{hovakimyan2010l1, wu2024l1quad,
huang2023datt}, disturbance observers~\cite{han2009adrc}, and robust or
learning-based model predictive control (MPC)~\cite{hewing2020learning, ostafew2016robust} reject
uncertainty or learned disturbance models while tracking a fixed
reference. Safety-constrained variants additionally certify constraint
satisfaction, but typically assume fixed disturbance
sets~\cite{langson2004robust, buch2021robust, iyengar2005robust,
abdul2025nonlinear} or distributions~\cite{cannon2009probabilistic,
schmid2025computing, chen2024probabilistic, saravanos2024distributed,
cosner2023robust}\revdel{, e.g., in robust and stochastic MPC, control barrier
functions (CBFs), dynamic programming, and safe RL}. Leveraging online
data to estimate or anticipate disturbances reduces this
conservatism~\cite{zintgraf2020varibad, rakelly2019efficient,
wu2021external, oshin2024robust, wang2025seal} and enables probabilistic
safety certificates~\cite{berkenkamp2017safe,hewing2019cautious, gao2025learning,
lapandic2024metampc, schmid2022real, zuliani2025closed}. However, these
approaches \revdel{remain tied to a given reference: they }improve tracking and
local safety, but cannot \revdel{improve performance by replanning the reference
itself}\revadd{replan the reference} when it proves infeasible or maintain real-time solvability subject to cluttered reach-avoid constraints.

\subsection{Robust Control Hierarchies}
\revdel{In contrast with the previous class, which adapts the low-level control
to reject disturbances while tracking a fixed reference, hierarchical}\revadd{Hierarchical}
architectures adapt the reference signal itself based on disturbance
data~\cite{singh2017robust, coursey2025rl, wang2025hybrid}. However, these works typically assume fixed disturbance
sets~\cite{kogel2025safe, herbert2017fastrack, fan2020fast,
stamouli2026layered, kolmanovsky2014reference, seo2019robust} or
distributions~\cite{schmid2026maximizing}, while others heuristically
replan references for the full reach-avoid problem, impeding real-time applicability~\cite{romero2022timeoptimal,
zhou2024internal, otte2016anytime, zhang2024change}. In contrast, BRACE frequently outputs only short-horizon reference paths for a lower-level Linear Quadratic Regulator (LQR) that are tracked sequentially and chosen to mitigate local disturbance estimates.

% In contrast with the previous class of methods, which adapt the low-level
% control to reject disturbances while tracking a fixed reference, hierarchical
% architectures instead adapt the reference signal itself based on disturbance
% data. Safety certificates for such hierarchical control architectures have recently gained increasing attention. However, such works typically assume fixed disturbance sets \cite{kogel2025safe, herbert2017fastrack, fan2020fast, stamouli2026layered, kolmanovsky2014reference} and distributions \cite{schmid2026maximizing}, while others heuristically replan online to react to unmodelled perturbations \cite{romero2022timeoptimal, zhou2024internal, otte2016anytime, zhang2024change}. Unlike these approaches, BRACE avoids ad-hoc replanning, which is often intractable in real-time. Instead, our method synthesizes an adaptive IMM-belief-based residual policy that manipulates the nominal reference signal to maximize the system's safety.
\subsection{Residual Policy Learning}
To render BRACE adaptive to online disturbance estimates and efficient to train, we apply the concept of residual policies, i.e., improving existing controllers by
learning an additive correction on top of it~\cite{johannink2019residual, ahmed2023rl,yuan2025policy}.
This formulation is commonly used when a reasonable nominal behavior (a disturbance-free reference path) is
available, and the residual only needs to correct remaining errors arising from problem variations (perturbations). We further improve training performance over methods from the literature by introducing a heuristic that enables to train the nominal and residual policy simultaneously rather than sequentially.

%%%%%%%%%%%%%%%%%%%%%%%%%%%%%%%%%%%%%%%%%%%%%%%%%%%%%%%%%%%%%%%%%%%%%%%%%%%%%%%%
\section{Problem Formulation}
\label{sec:problem}
\revdel{We consider a system with state $x_t\in\mathcal{S}$, position $p_t$, and
control input $u_t$, with nominal
one-step model $\hat{x}_{t+1}=f_0(x_t,u_t)$. The realized transition is
affected by a local disturbance as follows
\begin{equation}
    x_{t+1}=f_0(x_t,u_t)+B_d d_t,\qquad d_t=W(p_t),
\end{equation}
where $W$ is an unknown spatial disturbance field. The disturbance is
inferred from the prediction error $B_d d_t = x_{t+1}-\hat{x}_{t+1}$ and estimated by the IMM.

 We adapt references tracked by a lower-level controller online using the IMM disturbance belief. Our goal is to adapt the reference to maximize the probability of reaching a goal set $\mathcal{G}$ before entering the obstacle set $\mathcal{O}$.
}
\revadd{We consider a system with state $x_t\in\mathcal{S}$ (including a position $p_t$), and
control input $u_t$, with nominal
one-step model $\hat{x}_{t+1}=f_0(x_t,u_t)$. The system is subject to an additive perturbation
\begin{equation}
    x_{t+1}=f_0(x_t,u_t)+B_d W(x_t),
    \label{eq_dynamics}
\end{equation}
where $W\sim P_w$ is an unknown spatial disturbance field sampled from a probability law $P_w$. The controller observes neither $W$ nor $P_w$. In simulation,
we sample disturbance fields from $P_w$ both to train and to evaluate
the policy. Our goal is to maximize the probability of reaching a goal set $\mathcal{G}$ before entering the obstacle set $\mathcal{O}$.}

%-----------------------------------------------------------------------

%%%%%%%%%%%%%%%%%%%%%%%%%%%%%%%%%%%%%%%%%%%%%%%%%%%%%%%%%%%%%%%%%%%%%%%%%%%%%%%%
% \section{Disturbance-Aware Safety Analysis}
% \label{sec:safety}

% We characterize BRACE safety in two steps. First, we establish how
% bounded actuation and disturbance define an intrinsic feasibility
% geometry, motivating planning-level adaptation
% (Section~\ref{sec:margin}). 
% Second, within this planning architecture, we quantify safety under bounded belief error (Section~\ref{sec:belief_bound}).
% \subsection{Disturbance-Aware CBF Feasibility Margin}
% \label{sec:margin}
\begin{figure}[t]
    \centering
    \includegraphics[width=0.9\linewidth]{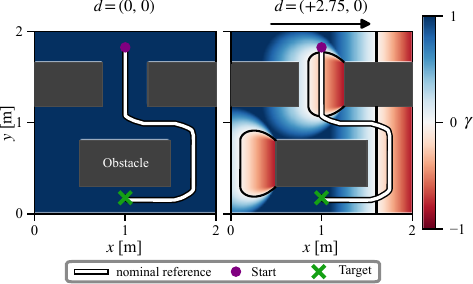}
    \vspace{-0.2cm}
    \caption{\revdel{CBF feasibility margin $\gamma(p,d)$~\eqref{eq:gamma} over the
workspace, for a single-integrator instantiation of \eqref{eq_dynamics}
($f=0$, $g=B_d=I$, $\alpha(h)=k_0 h$, $u_{\max}=2$, $k_0=2$) with $h$ a
signed-distance function for the obstacles, giving
$\gamma(p,d)=u_{\max}\|\nabla h(p)\|-\nabla h(p)^{\top}d-k_0 h(p)$.
For each fixed $d$, this is a field over the 2D workspace
(blue: $\gamma>0$, feasible; red: $\gamma<0$, infeasible; black curve:
$\gamma=0$). Left: $d=0$, the entire free space is feasible.
Right: $d=(2.75,0)$, past the threshold $d^\star$, infeasible pockets
appear downwind of the obstacles while the geometry is unchanged.
The white curve is the clearance-maximizing reference, identical in both
panels; under disturbance it crosses an infeasible region (right), where
the chosen CBF condition is locally infeasible under the input bound.}\revadd{Illustrative local CBF feasibility margin~\eqref{eq:gamma} for
$\dot p=u+d$, $\|u\|_2\leq2$, and $\alpha(h)=2h$, using the obstacle
signed-distance field. Blue/red indicates positive/negative margin; black
marks zero. The unchanged white reference crosses a negative-margin region
under $d=(2.75,0)$ (right), unlike $d=0$ (left). }}
    \label{fig:cbf-margin}
    \vspace{-0.4cm}
  \end{figure}
  
\section{Motivation for Planning-Layer Adaptation}
\label{sec:motivation}
\revdel{Before introducing BRACE, we formalize why disturbance adaptation should
occur at the planning layer.} Under bounded actuation, \revdel{a disturbance does
not simply create tracking error; it may make it infeasible to remain safe.}\revadd{disturbances can limit a tracking controller's ability to maintain safety along a prescribed reference, motivating adaptation of the reference itself.}
% The feasibility discussion is moved below the proposition, after Fig. 2.
\revdel{The feasibility of remaining safe depends on the dynamics, safe set,
disturbance direction, and actuation limit, and therefore varies across the
state space. While a tracking controller can follow
a prescribed reference more tightly with increased actuation, it cannot change the reference itself. A
planner can instead route the reference through higher-safety regions
when disturbance information is available.}
We \revdel{formalize this through the theory of CBFs.}\revadd{illustrate this limitation through the local feasibility of a control barrier function (CBF) condition.} 

Consider the control-affine system
\begin{equation}
\dot{x} = f(x) + g(x)u + B_d d,
\quad
u \in \mathcal{U} := \{u : \|u\| \le u_{\max}\},
\label{eq_dynamics_motivate}
\end{equation}
with safe set $\mathcal{C} := \{x : h(x) \le 0\}$ and extended class-$\mathcal{K}$
function $\alpha$. The system remains within $\mathcal{C}$ if $\dot{h}(x) \leq -\alpha(h(x))$ \cite{ames2016control}. \revdel{Inserting the dynamics \eqref{eq_dynamics_motivate} and rearranging yields
\begin{equation}
    0 \leq -L_g h(x) u- L_f h(x)  - \nabla h(x)^\top B_d d-\alpha(h(x)).
\end{equation}}
The CBF feasibility margin at state $x$ under
disturbance $d$ is the maximum value of \revdel{the right-hand side}\revadd{$-\dot h(x)-\alpha(h(x))$} over $u \in \mathcal{U}$, \revadd{which is}
\begin{equation}
\begin{aligned}
\gamma(x,d) :=\; & \|L_g h(x)\|\,u_{\max} - L_f h(x) \\
                 & - (\nabla h(x))^\top B_d\, d - \alpha(h(x)).
\end{aligned}
\label{eq:gamma}
\end{equation}
The chosen CBF condition is locally feasible at $x$ if and only if
$\gamma(x,d)\geq 0$.

\begin{proposition}[Disturbance-dependent CBF feasibility
threshold]
\label{prop:margin}
Assume the nominal CBF condition is feasible on $\mathcal{C}$, i.e.,
\begin{equation}
\revdel{\|L_g h(x)\|\,u_{\max} - L_f h(x) - \alpha(h(x))}\revadd{\gamma(x,0)} \ge 0,
\quad \forall x \in \mathcal{C}.
\label{eq:nominal_feasible}
\end{equation}
Let
\revdel{\begin{equation}
\mathcal{C}_+ :=
\left\{x \in \mathcal{C} :
\|(\nabla h(x))^\top B_d\| > 0 \right\},
\end{equation}}\revadd{$\mathcal{C}_+ := \{x \in \mathcal{C} : \|(\nabla h(x))^\top B_d\| > 0\}$,}
and suppose $\mathcal{C}_+ \neq \emptyset$. \revdel{Assuming a minimizer exists, let}\revadd{Define}
\begin{equation}
d^\star :=
\min_{x \in \mathcal{C}_+}
\frac{\begin{gathered}\revdel{\|L_g h(x)\|\,u_{\max} - L_f h(x) - \alpha(h(x))}\ifshowrevision\\\fi\revadd{\gamma(x,0)}\end{gathered}}
     {\|(\nabla h(x))^\top B_d\|}\geq 0\revdel{.}\revadd{,}
\label{eq:dstar}
\end{equation}
\revadd{where the minimum is assumed to be attained at some $x^\star\in\mathcal{C}_+$.}
Then,
\begin{enumerate}
\item[(i)] for every disturbance satisfying $\|d\| \leq d^\star$,
the \revadd{selected }CBF condition remains feasible at every $x \in \mathcal{C}$, i.e.,
$\gamma(x,d) \ge 0$;
\item[(ii)] for every disturbance magnitude $\bar d > d^\star$, there exist a
disturbance vector $d$ with $\|d\| = \bar d$ and a state $x \in \mathcal{C}_+$
such that $\gamma(x,d) < 0$. \revdel{At this state, the chosen disturbance-aware CBF condition
cannot be satisfied by any admissible control $u \in \mathcal{U}$.}
\end{enumerate}
\end{proposition}

\begin{proof}
We first prove (i). Let $x\in\mathcal{C}_+$. \revdel{Then,
\begin{align*}
    d^\star &\leq \frac{\|L_g h(x)\|\,u_{\max} - L_f h(x) - \alpha(h(x))}
     {\|(\nabla h(x))^\top B_d\|}.
 \end{align*}
 Hence,
 \begin{align*}
    0&\leq \|L_g h(x)\|\,u_{\max}\!-\!L_f h(x)\!-\!\alpha(h(x))\!-\!d^\star\|(\nabla h(x))^\top B_d\| \\
     &\leq \|L_g h(x)\|\,u_{\max}\!-\!L_f h(x)\!-\!\alpha(h(x))\!-\!(\nabla h(x))^\top B_d\, d,
\end{align*}
where the second inequality uses
$(\nabla h(x))^\top B_d\, d \le \|(\nabla h(x))^\top B_d\|\,\|d\|
\le d^\star\|(\nabla h(x))^\top B_d\|$ and $\|d\|\le d^\star$.
}\revadd{By Cauchy--Schwarz and \eqref{eq:dstar},
$\gamma(x,d)\geq\gamma(x,0)-\|(\nabla h(x))^\top B_d\|\,\|d\|\geq0$
for $\|d\|\leq d^\star$.}
\revdel{This means $\gamma(x,d)\geq 0$. Further, any $x\in\mathcal{C}\setminus\mathcal{C}_+$ yields $\|(\nabla h(x))^\top B_d\| = 0$. Then, $\gamma(x,d)\geq 0$ holds by the assumption that the CBF condition \eqref{eq:nominal_feasible} is nominally feasible on $\mathcal{C}$.}
\revadd{For $x\in\mathcal{C}\setminus\mathcal{C}_+$, the disturbance term vanishes, so $\gamma(x,d)=\gamma(x,0)\geq0$ by \eqref{eq:nominal_feasible}.}

For (ii), let $\bar d > d^\star$. Since the minimum in \eqref{eq:dstar} is
attained, there is $x^\star \in \mathcal{C}_+$ with
$\|(\nabla h(x^\star))^\top B_d\|\, d^\star = \revdel{\|L_g h(x^\star)\| u_{\max} - L_f h(x^\star) - \alpha(h(x^\star))}\revadd{\gamma(x^\star,0)}$. Choose
\[
d = \bar d\,\frac{B_d^\top \nabla h(x^\star)}{\|B_d^\top \nabla h(x^\star)\|},
\]
so that $\|d\| = \bar d$\revdel{ and
$(\nabla h(x^\star))^\top B_d\, d = \|(\nabla h(x^\star))^\top B_d\|\,\bar d$}.
Then $\gamma(x^\star, d) = \|(\nabla h(x^\star))^\top B_d\|\,(d^\star - \bar d) < 0$\revdel{, since $\bar d > d^\star$ and $\|(\nabla h(x^\star))^\top B_d\| > 0$}. At
$x^\star$, no admissible control satisfies the CBF condition.
\end{proof}

\begin{remark}[Implication for planning-layer adaptation]
\label{rem:arch}
The threshold $d^\star$ depends only on
$(f,g,B_d,h,\alpha,u_{\max})$, not on a
specific controller. Thus, under bounded actuation, disturbance direction and magnitude induce a spatial feasibility margin over the safe set. \revdel{A tracking controller spends control authority to reduce error
relative to a prescribed reference, but cannot change the geometry of
that reference. A planning-layer policy can instead choose references
that pass through higher-margin regions. BRACE achieves that
by making the reference depend on an online
disturbance belief through a residual policy.}
\end{remark}
Figure~\ref{fig:cbf-margin} makes the above observation more concrete.
\revadd{The feasibility margin varies across the state space with the dynamics, safe set, disturbance, and actuation limit. Whereas a tracker follows a prescribed reference, a planner can use disturbance information to route through higher-margin regions.}
\revdel{The white curve is the
clearance-maximizing reference, identical in both panels. At $d=(0,0)$ it lies
entirely in the feasible region; at $d=(2.75,0)$ it crosses an infeasible
pocket ($\gamma<0$) that has opened downwind of an obstacle, where the chosen CBF condition is locally infeasible under the input bound, exactly as in Theorem~\ref{prop:margin}(ii). The same
reference is safe without disturbance and unsafe with it, and because the
pocket's location depends on the disturbance, which is unknown when the
reference is generated, the reference must be adapted online rather than fixed
in advance.}

\section{BRACE: Belief-Guided Residual Adaptive Control}
\label{sec:method}
\revdel{At deployment, BRACE comprises an IMM disturbance belief estimator
(Section~\ref{sec:imm}), a belief-guided residual policy
(Section~\ref{sec:residual}), and a low-level Linear Quadratic Regulator (LQR), see
Section~\ref{sec:lqr}. }
\revadd{Motivated by Remark~\ref{rem:arch}, we introduce the BRACE algorithm that adapts references to online disturbance estimates. Instead of planning a full reference to the goal, the reference generator outputs a short-horizon trajectory that is replanned at every decision step, which keeps computation low and allows the reference to adapt as the disturbance belief evolves. The reference generator is separated into a nominal and disturbance-adaptive residual policy. This enables the nominal policy to solve potentially complex reach-avoid objectives, while a smaller neural network adapts to disturbances.} Both policies are trained with a CBF-\revdel{based}\revadd{inspired} safe
RL objective (Section~\ref{sec:dcbf}). While it is common practice to sequentially train first the nominal, then the residual policy, we enable simultaneous training through a likelihood-ratio advantage
reweighting for credit assignment (Section~\ref{sec:reweight}). For clarity of presentation, we describe our algorithm using a quadcopter as a running example.

\subsection{IMM Disturbance Belief Estimator}
\label{sec:imm}
\revdel{
At every time step, we compute the position prediction error
\begin{equation}
    e_{t+1} = p_{t+1} - \hat{p}_{t+1},
\end{equation}
where $p_{t+1}$ is the realized planar position and $\hat{p}_{t+1}$ is the
position predicted by the nominal model $f_0(x_t, u_t)$. This error captures
the mismatch between the nominal prediction and the realized closed-loop
motion, and serves as an indirect measurement of the external disturbance.

The IMM maintains two Gaussian disturbance models with different process
noise levels, $Q^{(slow)} \ll Q^{(fast)}$. The low-noise model represents
slowly varying disturbances, while the high-noise model reacts more
quickly to abrupt disturbance changes. At each step, the IMM updates both
the disturbance estimate and the posterior probability of each model
based on how well that model explains the observed tracking error.

The resulting belief is summarized as
\begin{equation}
    b_t =
    (\mu_x,\mu_y,\sigma_x,\sigma_y,p_\mathrm{fast}),
\end{equation}
where $(\mu_x,\mu_y)$ is the estimated disturbance mean,
$(\sigma_x,\sigma_y)$ represents estimation uncertainty in each
disturbance direction, and $p_\mathrm{fast}$ is the posterior probability
of the fast-changing model. Thus, $\sigma_x,\sigma_y$ quantify uncertainty
in the disturbance estimate, while $p_\mathrm{fast}$ captures uncertainty
in the disturbance model choice.
}
\revadd{From Eq.~\eqref{eq_dynamics} we obtain $z_t\coloneqq B_dW(x_t) = x_{t+1} - \hat{x}_{t+1}$. Estimating the whole disturbance field $W$ is generally challenging, particularly for high-dimensional state spaces and discontinuous disturbance fields. Instead, we use past measurements $(z_{t-1}, z_{t-2},\dots)$ to estimate the current disturbance $d_t$. We model the disturbance as a random walk $d_{t+1}={d}_t+q_t$, $q_t\sim \mathcal{N}(0,Q)$, observed through $z_{t} = d_t + v_{t}$, $v_{t}\sim\mathcal{N}(0,R)$, where $v_{t}$ accounts for sensor noise and model mismatch. A Kalman filter then yields a Gaussian posterior over $d_t$. Fixing $R\succ 0$, a large $Q$ allows for rapid changes of the disturbance estimate, while a small $Q$ smooths over previous measurements. To accommodate both in the face of an unknown wind field, we use two models with $Q^{(fast)}\succ Q^{(slow)}$ and allow switching between both models with predefined probabilities, leading to an Interacting Multiple Model (IMM) estimator~\cite{blom1988interacting}. The IMM also computes the posterior probability $p_\mathrm{fast}$ that the
fast model is active, with $p_\mathrm{slow}=1-p_\mathrm{fast}$.

For a quadcopter whose $x$- and $y$-positions are perturbed, we summarize the belief information as $b_t = (\mu_x,\mu_y,\sigma_x,\sigma_y,p_\mathrm{fast})\in\mathbb{R}^5$, where $(\mu_x,\mu_y)$ is the estimated mean perturbation on the respective coordinates and
$(\sigma_x,\sigma_y)$ the corresponding uncertainty. The process and measurement covariances, model-transition
probabilities, and initialization are listed in
Table~\ref{tab:implementation_settings}.}

\subsection{Belief-Guided Residual Policy}
\label{sec:residual}

\revadd{The reference-generator is implemented as a Gaussian policy, $\pi(\cdot | s_t) = \mathcal{N}(\,\cdot\,;\, \mu_t(s_t),\, \Sigma_t(s_t))$,
with $s_t = (o_t, b_t)$,
where the observation $o_t \in \mathbb{R}^8$ stacks the drone position, velocity,
goal position, and a clipped relative-position vector to the nearest obstacle
within a $0.5$\,m sensing radius and $b_t$ represents the belief from IMM filter.

In principle, the mean and covariance of the Gaussian policy can be learned directly through a neural network. However, reach-avoid navigation tasks are already hard to solve due to sparse rewards. Additionally, learning to adapt the reference to varying disturbance configurations and belief states enlarges the search space, which can make learning more challenging. 

To address this problem, we decompose $\mu_t$ into a nominal and a belief-guided residual term
\begin{equation}
    \mu_t(s_t)
    =
    \mu_{\mathrm{nom},t}(o_t)
    +
    \mu_{\mathrm{res},t}(o_t,b_t).
    \label{eq:full_action}
\end{equation} 
The residual factorizes as $\mu_{\mathrm{res},t}=\phi_{\mathrm{mean}}\,\phi_{\mathrm{scale}}$,
where $\phi_{\mathrm{mean}}$ maps the nominal action, drone state, goal, and
$(\mu_x,\mu_y)$ to a 2D correction, and $\phi_{\mathrm{scale}}$ maps
$(\sigma_x,\sigma_y,p_\mathrm{fast})$ to a scalar that modulates its magnitude. The
covariance $\Sigma_t(s_t)$ is produced by a separate network and shared by the nominal and full policies. We implement all functions,
$\mu_{\mathrm{nom}}$, $\phi_{\mathrm{mean}}$, $\phi_{\mathrm{scale}}$, and $\Sigma$ as multi-layer perceptrons (MLPs) trained jointly
under the PPO objective of Section~\ref{sec:dcbf}, and empirically demonstrate the benefit of the separation in Eq.~\eqref{eq:full_action} through numerical experiments in Section \ref{sec:sim}.}
\revdel{
The high-level policy is Gaussian,
\begin{equation}
    \pi_\theta(a_t \mid s_t) = \mathcal{N}(a_t;\, \mu_t,\, \Sigma_t),
    \qquad s_t = (o_t, b_t),
\end{equation}
whose mean is decomposed into a nominal and a belief-guided
residual:
\begin{equation}
    \mu_t
    =
    \mu_{\mathrm{nom},t}
    +
    \mu_{\mathrm{res},t}.
    \label{eq:full_action_past}
\end{equation}
Here, observation $o_t \in \mathbb{R}^8$ stacks the drone position, velocity,
goal position, and a clipped relative-position vector to the nearest obstacle
within a $0.5$\,m sensing radius and $b_t$ represents the belief from IMM filter. 
The nominal and residual components are defined as
\begin{equation}
\begin{aligned}
    \mu_{\mathrm{nom},t}
    &=
    \pi_{\mathrm{nom}}(o_t), \\
    \mu_{\mathrm{res},t}
    &=
    \phi_{\mathrm{mean}}
    \!\left(
        [\mu_{\mathrm{nom},t},\, o_t,\, \mu_x,\, \mu_y]
    \right)
    \,
    \phi_{\mathrm{scale}}
    \!\left(
        [\sigma_x,\, \sigma_y,\, p_{\mathrm{fast}}]
    \right).
\end{aligned}
\label{eq:residual_policy}
\end{equation}
Here, $\phi_{\mathrm{mean}}$ outputs a 2D correction driven by the disturbance
mean $(\mu_x,\mu_y)$, while $\phi_{\mathrm{scale}}$ outputs a scalar driven
only by the uncertainty features $(\sigma_x,\sigma_y,p_{\mathrm{fast}})$,
thereby modulating the residual magnitude uniformly across axes. The diagonal
covariance $\Sigma_t$ is produced by a separate network that also reads the
observation and belief, independent of the mean branches. The nominal branch
$\pi_{\mathrm{nom}}$, the residual branches $\phi_{\mathrm{mean}}$ and
$\phi_{\mathrm{scale}}$, and the covariance network are multi-layer perceptrons (MLPs) trained jointly
under the PPO objective of Section~\ref{sec:dcbf}.

}

% \subsection{Trajectory Generation and LQR Execution}
% \label{sec:lqr}
% The action $a_t$ is converted to a short-horizon kinematically feasible
% reference $\tau_t$. A time-varying LQR tracks $\tau_t$ with frequency
% $f_\mathrm{LQR}$, decoupling fast stabilization from
% adaptive trajectory selection.

\subsection{\revdel{Trajectory}\revadd{Reference} Generation and LQR Execution}
\label{sec:lqr}
At a rate of $f_\mathrm{plan} = 2$ Hz, the planner samples a velocity setpoint $a_t\sim\pi(\cdot|s_t)$ where $a_t\in\mathbb{R}^2$ and generates a short-horizon reference $\tau_t$ by constant-velocity extrapolation from the current measured position $p_t$. A finite-time LQR then tracks $\tau_t$ at $f_\mathrm{LQR} = 40$ Hz by outputting thrust and body-rate commands. This separates two timescales: the planner adapts the reference under the inferred disturbance, while the LQR regulates tracking errors against that reference.

\subsection{CBF-\revdel{Based}\revadd{Inspired} Safe Training}
\label{sec:dcbf}
We train the high-level \revdel{planner}\revadd{reference generator}
using the discrete-CBF (DCBF) PPO training framework of
Zhang et al.~\cite{zhang2025discrete}.
Let $h(x_t) = \delta - \mathrm{dist}(p_t)$ denote the safety signal, where $\mathrm{dist}(p_t)$ is the planar clearance from position $p_t$ to the nearest obstacle or arena boundary and $\delta$ is a fixed safety margin; $h(x_t) > 0$ thus indicates a constraint violation.
% \revadd{Let $h(x_t)$ denote the constraint function, where $h(x_t)\le 0$ is safe and $h(x_t)>0$ indicates the constraint violation.}
In addition
to the standard PPO reward critic, we learn a constraint value function
\revdel{$V^h_\phi(o_t)$ that tracks the worst future value
of $h$ along a trajectory.}\revadd{$V^h_t := V^h(o_t,b_t)$ that approximates the worst future
constraint violation along a trajectory, $\max_{k\geq0}h(x_{t+k})$.
We fit $V^h$ by least squares to targets computed from the values of $h$ along rollouts of the mean (deterministic) policy under sampled disturbance fields.}
\revadd{Rollout transitions include the intervening LQR execution (Section~\ref{sec:lqr}).}
The CBF residual
\revdel{
\begin{equation}
\begin{aligned}
C_t ={}& V^h_\phi(o_{t+1})
       - V^h_\phi(o_t) + \alpha_h V^h_\phi(o_t)
\end{aligned}
\end{equation}
}
\revadd{
\begin{equation}
\begin{aligned}
C_t ={}& V^h_{t+1} -V^h_t +\alpha_h V^h_t 
\end{aligned}
\end{equation}
}

with contraction rate $\alpha_h\in(0,1)$ is non-positive on transitions
satisfying the \revdel{CBF decrease}\revadd{learned DCBF} condition.
We \revadd{use this condition to }shape the reward advantage as
\begin{equation}
    \tilde{A}_t =
    \mathds{1}[C_t \le 0]\,A_t
    - w_h\, \max(C_t+\varepsilon_h,\,0),
    \label{eq:advantage}
\end{equation}
where $A_t$ is the normalized PPO reward advantage, $w_h > 0$ weights the safety objective, and $\varepsilon_h > 0$ is an
activation margin. \revadd{Continuous-time safety could further be established under Lipschitz bounds by enforcing tightened reference constraints~\cite{choi2026safe}.}
We refer readers to~\cite{zhang2025discrete} for full algorithmic details.
% Reward updates thus apply only when the CBF
% decrease condition holds ($C_t \le 0$); transitions that violate it
% instead contribute a penalty proportional to the margin.

\subsection{Likelihood-Ratio Advantage Reweighting}
\label{sec:reweight}
\begin{figure}[t]
    \centering
    \includegraphics[width=0.6\linewidth]{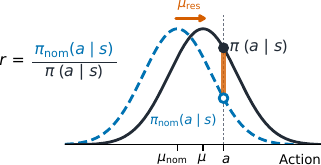}
    % \vspace{-0.2cm}
    \caption{Likelihood-Ratio Advantage Reweighting. The likelihood ratio $r$
in~\eqref{eq:reweight_ratio} compares the sampled action's likelihood
without and with the residual policy.}
\vspace{-0.4cm}
    \label{fig:likelihood_ratio}
\end{figure}
The CBF-\revdel{based}\revadd{inspired} PPO objective above produces
a safety-shaped advantage $\tilde A_t$ for the combined policy.
However, in BRACE the mean action is the sum of two branches, $\mu_{\mathrm{nom}}$ and $\mu_{\mathrm{res}}$. Applying
the same advantage to both
\revdel{misattributes}\revadd{can misattribute} credit in either direction. A high-reward rollout with a near-zero residual still updates the
residual branch and can drive it to imitate the nominal action, making
it redundant; conversely, a rollout that succeeded through a large
residual correction still updates the nominal branch and can pull it
toward disturbance-specific behavior. As a remedy, we use a PPO-specific
\revadd{heuristic} credit assignment rule
\revdel{based on how much each branch contributes to the likelihood
of the sampled action.}\revadd{based on the sampled action's relative
likelihood under the full and nominal-only policies.}

Let $\pi(a\mid s)$ be the
\revdel{Gaussian rollout}\revadd{current full} policy
with mean $\mu$ as in Eq.~\eqref{eq:full_action}, and let $\pi_\mathrm{nom}(a\mid s)$ be the \revadd{nominal-only} policy where $\mu_\mathrm{res}$ is set to zero.
We define the likelihood ratio \revadd{(Fig.~\ref{fig:likelihood_ratio})} and the branch-specific
credit weights as
\begin{equation}
    r(s,a) = \frac{\pi_\mathrm{nom}(a\mid s)}
                  {\pi(a\mid s)},
    \label{eq:reweight_ratio}
\end{equation}
\begin{equation}
\begin{aligned}
w^\mathrm{nom}
    &= \revadd{\mathrm{clip}(r(s,a),0,1)},\\
w^\mathrm{res}
    &= \revadd{\mathrm{clip}(1-r(s,a),0,1)},
\end{aligned}
\label{eq:reweight_weights}
\end{equation}
where we assume that $\Sigma_t\succ 0$ for all $t$. We then \revdel{split}\revadd{weight} the safety-shaped advantage
from~\eqref{eq:advantage} as $\tilde A_t^i = w^i\tilde A_t$ \revadd{for}
$i\in\{\mathrm{nom},\mathrm{res}\}$. \revdel{Since $w^\mathrm{nom}$ tracks $r$ and $w^\mathrm{res}$ tracks $1-r$, the weights follow the effect of the residual on the sampled
action: when the residual has little effect, the nominal and full
policies assign similar likelihoods, $r(s,a)\approx1$, and the nominal
branch receives most of the credit; conversely, when the residual
makes the sampled action far more likely under the full policy,
$r(s,a)\to0$, and the residual branch receives most of the credit. The nominal and residual branches are then updated with
$\tilde A^\mathrm{nom}_t$ and $\tilde A^\mathrm{res}_t$, respectively.}\revadd{The nominal branch, together with the shared covariance $\Sigma_t$, is updated with $\tilde A^\mathrm{nom}_t$, and the residual branch with $\tilde A^\mathrm{res}_t$.
We lower-bound both weights at $0.05$ so that neither branch stops
learning.}
\revdel{This likelihood-based credit assignment encourages each branch to
learn primarily from the transitions it better explains, while the
floor $\varepsilon_w$ prevents either branch from
losing the gradient entirely.}
% \revadd{For a sampled action, the residual branch receives greater
% weight when the full policy assigns substantially higher density
% than the nominal-only counterfactual ($r\ll1$).
% When the nominal counterfactual assigns equal or higher density
% ($r\geq1$), the residual weight is zero before flooring.

\revadd{On average, the weighting rule assigns more weight to the nominal
branch when the two action distributions largely overlap, and
increasingly weights the residual branch as the distributions
become more distinguishable. 

% To quantify this relationship, let
% $p=\pi_\mathrm{full}(\cdot\mid s)$ and
% $q=\pi_\mathrm{nom}(\cdot\mid s)$.
% For actions sampled from the current policy, $a\sim p$, at a fixed input, the residual
% weight $w^\mathrm{res}=\max(1-q/p,0)$ satisfies
% \begin{equation}
% \mathbb E_p[w^\mathrm{res}]
%     = \operatorname{TV}(p,q)=2\Phi(D/2)-1.
% \label{eq:reweight_tv}
% \end{equation}
% Here $\operatorname{TV}(p,q)=\tfrac12\int|p-q|\,da$,
% $\Phi$ is the standard normal CDF, and
% $D^2=\delta^\top\Sigma^{-1}\delta$, with gated displacement
% $\delta=\mu_\mathrm{full}-\mu_\mathrm{nom}$ and shared Gaussian
% action covariance $\Sigma$. Thus, expected residual weight grows
% with standardized displacement; expected nominal weight is its complement.
}

% \revadd{With the credit weights held fixed during differentiation,
% the nominal-weighted PPO loss updates the nominal and covariance
% networks, while the residual-weighted loss updates only the
% residual and gate networks. The covariance network also receives
% the entropy update. We apply
% $w^i\leftarrow\max(w^i,\varepsilon_w)$ with $\varepsilon_w=0.05$
% without renormalization to prevent zero weighting of branch updates.\color{red} I do not understand this paragraph\color{black}}

\subsection{\revdel{A-Posteriori Safety Validation}\revadd{A-Posteriori Reach-Avoid Performance Estimation}}
\label{sec:belief_bound}
\revadd{BRACE does not provide a-priori closed-loop safety guarantees. We therefore evaluate its reach-avoid performance a-posteriori using Monte-Carlo simulations and report lower confidence bounds on the success probability.}\revdel{We complement training with an a-posteriori statistical validation of
the closed-loop hierarchy.} Suppose that the disturbance fields are
sampled from a distribution $P_w$, and we run $N$ Monte-Carlo simulations with $W\sim P_w$. If our control hierarchy satisfies the reach-avoid specification in $m$ out of $N$ trials in simulation, then a one-sided Clopper--Pearson lower confidence bound on
the success probability under $P_w$ is
\begin{equation}
    p\geq I^{-1}_x\left(\revdel{1-\beta}\revadd{\beta}; m, N - m + 1\right),
\end{equation}
with confidence \revdel{$\beta\in(0,1)$, where}\revadd{$1-\beta$, where $\beta\in(0,1)$ is the tail probability (we use $\beta=0.05$) and} $I^{-1}_x$ is the inverse
cumulative distribution function of the beta distribution
\cite{agresti2013categorical}. 

\section{Simulation Experiments}
\label{sec:sim}

\begin{figure}[t]
    \centering
    \includegraphics[width=\linewidth]{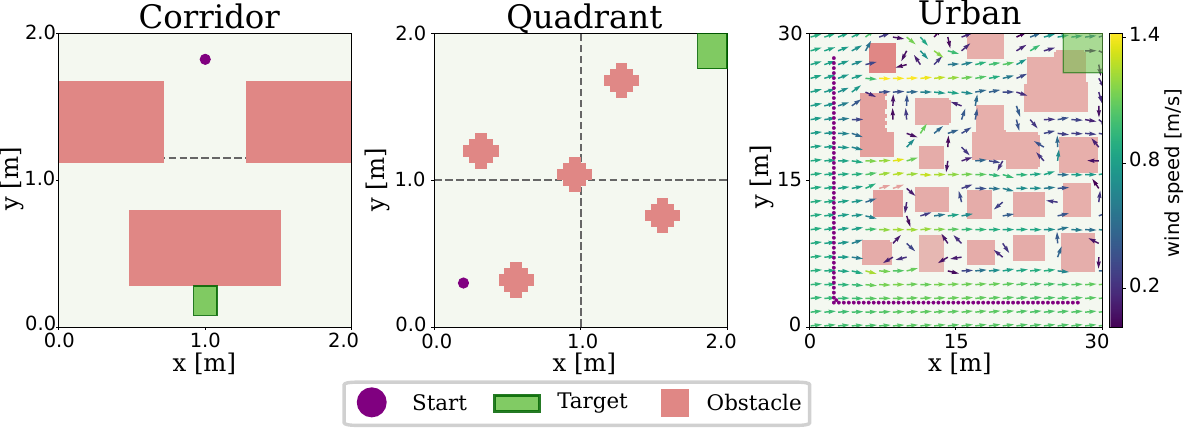}
    \vspace{-0.5cm}
    \caption{Simulation environments. \texttt{Corridor} is divided
    into two regions and \texttt{Quadrant} into four square regions,
    each with a piecewise-constant disturbance field. \texttt{Urban}
    uses a spatially varying wind field derived from measurement data.}
    \vspace{-0.4cm}
    \label{fig:simulation-environments}
\end{figure}

\subsection{Setup}
\label{sec:setup}
We evaluate simulated Crazyflie (27\,g) reach-avoid navigation in three
environments (\Cref{fig:simulation-environments}). \texttt{Corridor} and \texttt{Quadrant} use synthetic force fields, while \texttt{Urban} uses a realistic
urban wind field derived from measurement
data~\cite{so2024solving}. Episodes terminate on goal reaching,
collision, or timeout.

\texttt{Corridor} and \texttt{Quadrant} divide the workspace into spatial
zones, each applying a horizontal force with per-zone magnitude sampled
\revadd{independently and }uniformly in $[0.1, 0.5]$ of the drone weight
($0.027$--$0.132$\,N). At every LQR control step, this per-zone magnitude is
further perturbed by a stochastic scaling factor drawn uniformly from
$[0.9, 1.1]$ ($\pm 10\%$) to capture unmodeled gust variability.
\texttt{Corridor} uses two zones with disturbances along $\pm x$\revadd{, with
equal probability}; \texttt{Quadrant} uses four zones with disturbances in
arbitrary planar directions\revadd{, uniform in angle}. \revadd{This sampling law is the $P_w$ of
Section~\ref{sec:belief_bound}; the narrow condition instead draws magnitudes
from $[0.2,0.4]$, a different law rather than a restriction of the same one.}

\revadd{Table~\ref{tab:implementation_settings} summarizes the numerical
settings. The LQR uses linearized quadrotor dynamics with position,
velocity, and Euler-angle states.}

\begin{table}[t]
\centering
\caption{\revadd{Implementation settings. S: Corridor/Quadrant; U: Urban.
IMM covariances are in m$^2$.}}
\label{tab:implementation_settings}
\revadd{%
\footnotesize
\setlength{\tabcolsep}{2pt}
\setlength{\aboverulesep}{1pt}
\setlength{\belowrulesep}{1pt}
\renewcommand{\arraystretch}{1.0}
\begin{tabular}{@{}>{\raggedright\arraybackslash}p{0.11\columnwidth}
                  >{\raggedright\arraybackslash}p{0.19\columnwidth}
                  >{\raggedright\arraybackslash}p{0.66\columnwidth}@{}}
\toprule
Module & Parameter & Value \\
\midrule
IMM & Process noise & $Q^{(slow)}=10^{-6}I_2$, $Q^{(fast)}=10^{-3}I_2$ \\
 & Measurement & $R_{\rm IMM}=10^{-4}I_2$ \\
 & Transitions & $P=\bigl[\begin{smallmatrix}0.98&0.02\\0.05&0.95\end{smallmatrix}\bigr]$ (slow/fast order) \\
 & Initialization & Means $0$; covariances $10^{-4}I_2$; mode probabilities $(0.9,0.1)$ \\
\midrule
LQR & Horizon & $0.5$\,s (20 steps) \\
 & Weights & $Q=Q_f=\operatorname{diag}(4I_3,2I_3,I_3)$; $R_{\rm LQR}=0.01I_4$ \\
 & Limits & Normalized commands $[-1,1]$; attitude targets $\pm25^\circ$ \\
\midrule
DCBF 
& Margin $\delta$ & $0.02$\,m (S); $0.05$\,m (U) \\
 % & $\alpha_h$ & $0.5$ \\
  & $\alpha_h$, $\varepsilon_h$ & $0.5$, $0.1$ \\
 & Penalty $w_h$ & Initially $0.5$ (S), $0.01$ (U); $\times2$ at 50\% and again at 75\% of training \\
\midrule
PPO & Training & 512 parallel environments; 5000 updates \\
 & Rollouts & 30 steps (S); 40 (U) \\
 & PPO steps & $76.8$M (S); $102.4$M (U) \\
 & LR / clip & $5\times10^{-5}$ / $0.2$ \\
 & Discount & $0.99$ \\
\midrule
RP-MPC & Horizon & $3$\,s; replanning at $2$\,Hz \\
 & Limits & $|v_x|,|v_y|\leq0.5$\,m/s \\
 & Solver & IPOPT; max.\ 200 iterations; tol.\ $10^{-4}$ \\
\midrule
BRACE & $\mu_{\rm nom}$ & 2 hidden layers, 256 units each \\
 & $\phi_{\rm mean}$ & 2 hidden layers, 32 units each \\
 & $\phi_{\rm scale}$ & 1 hidden layer, 32 units \\
\bottomrule
\end{tabular}}
\vspace{-0.4cm}
\end{table}

The reward is $r = 10\cdot\mathbf{1}_\mathrm{target}$ for \texttt{Corridor}/\texttt{Quadrant}. \texttt{Urban} applies
relative-velocity drag from a fixed 30\,m $\times$ 30\,m wind field
(peak $\approx 0.18$\,N), with the reward augmented by a progress term
(weight $0.1$) and action penalty ($10^{-4}$). 

\revdel{All policies train with 512 parallel environments for 5000 PPO updates
(30-step rollouts, $76.8$M steps for \texttt{Corridor}/\texttt{Quadrant};
40-step rollouts, $102.4$M steps for \texttt{Urban}), using learning rate
$5\times10^{-5}$, PPO clip $0.2$, and discount
$0.99$. The nominal and residual policies are
$2\times256$ and $2\times32$ MLPs.}
\revdel{For each method and condition, we train 5 independent policies from different
random seeds and report mean $\pm$ std across them; each method is evaluated on
500 episodes for the synthetic environments and on a deterministic 100-point
start grid with a fixed goal for \texttt{Urban}. Classical planners,
which require no training, are evaluated on the same episodes.}\revadd{ All policies trained from scratch are first trained without wind for $500$ of $5{,}000$ PPO iterations (warmup), after which disturbance fields are sampled from $P_w$; Frozen-nominal trains only its residual, with wind from the start.

We separately quantify the variation due to training and the variation due to the sampled test fields. To capture the former, we train each learned method with five seeds, each keeping its best checkpoint on separate validation episodes; to capture the latter, for each wind condition we test every policy and planner on the same five batches of $500$ episodes, each generated with a different random seed ($2{,}500$ episodes in total). Table~\ref{tab:quad_main} reports the best-validation seed of each learned method, with the standard deviation across the five test batches; Table~\ref{tab:seed_lb} reports the variation across the five training seeds. }\texttt{Urban} uses a fixed goal and a deterministic
100-point start grid (mean $\pm$ std over seeds).

\subsection{Baselines and Ablations}
We compare against \revdel{tracking-level}\revadd{tracking- and planning-level} baselines and run \revdel{a
controlled ablation isolating the effect of each architectural choice}\revadd{ablations of belief conditioning, reweighting, and joint training}. \revdel{Both tracking baselines follow the same reference path: a centerline through the free space that is locally as far as possible from all obstacles, known as the medial axis~\cite{lavalle2006planning}.}\revadd{Both tracking baselines follow the same medial-axis reference, locally maximizing obstacle clearance~\cite{lavalle2006planning}.}
\begin{itemize}
    \item \textit{Centerline}: tracking of the medial-axis path, with no disturbance adaptation.
    \item \textit{DA-MPC}: disturbance-aware MPC,
          implemented as a quadratic programming (QP)-based
          linear MPC tracking the medial-axis reference path.
          DA-MPC uses the estimated disturbance mean $(\mu_x,\mu_y)$ from the IMM filter,
injected as an additive position-level feedforward in the MPC dynamics. The
feedforward gain is grid-searched over $1$--$40$ on held-out validation episodes.\revadd{ The regenerated results use gains of $8$ on \texttt{Quadrant} and $16$ on \texttt{Corridor}.}
    \item \revadd{\textit{RP-MPC}: a nonlinear replanning MPC
    baseline inspired by the disturbance-aware planning approach
    of Wu et al.~\cite{wu2021external}. It adds the IMM disturbance mean, multiplied by a drift scale, as a position drift in its prediction and optimizes velocity references through LQR, with inflated-obstacle constraints. A held-out sweep selects drift scale $4$, guidance weight $30$, and obstacle margin $0.04$\,m, with a fixed world margin of $0.08$\,m on both maps.}
    \item \textit{No-belief}: \revdel{BRACE with belief removed from
          observations.}\revadd{an RL reference generator without belief information.} Isolates the value of belief in trajectory
          generation.
    \item \textit{No-residual}: directly approximates $\mu_t$ in Eq.~\eqref{eq:full_action} with a single neural network. Isolates the
          contribution of the residual architecture.
    \item \textit{No-reweight}: belief-guided residual policy without
          advantage reweighting. Isolates the contribution of
          likelihood-ratio credit assignment.
    \item \textit{Frozen-nominal}: conventional residual training, where the
      nominal is pretrained for 3000 updates, then frozen while the residual is
      trained for 2000 more (5000 total, matching BRACE's budget). \revdel{Isolates the
      contribution of joint nominal--residual training.}
    \item \textit{BRACE-nominal-only}: full BRACE with the residual
          zeroed at inference. Isolates the residual's contribution at
          deployment.
    \item \textit{BRACE}: our full method.
\end{itemize}

\subsection{Synthetic Wind Results}

\begin{table}[t]
\centering
\caption{\revdel{Success rate (\%) on \texttt{Quadrant} and \texttt{Corridor}
reach-avoid. Mean $\pm$ std over 5 seeds $\times$ 500 episodes.}
\revadd{Success rate (\%; subscript: std across the five $500$-episode batches) and $95\%$ lower bound $L$; learned methods use their best-validation seed.}}
\label{tab:quad_main}
\label{tab:sim_ci}
% Original submitted table retained as a deletion; merged revision follows.
\revdel{\resizebox{\columnwidth}{!}{%
\begin{tabular}{lcccc}
\toprule
 & \multicolumn{2}{c}{\texttt{Quadrant}}
 & \multicolumn{2}{c}{\texttt{Corridor}} \\
\cmidrule(lr){2-3} \cmidrule(lr){4-5}
Method & $[0.1,0.5]$ & $[0.2,0.4]$ & $[0.1,0.5]$ & $[0.2,0.4]$ \\
\midrule
Centerline           & $10.7 \pm 1.2$ & $9.0 \pm 1.0$   & $0.0 \pm 0.0$  & $0.0 \pm 0.0$  \\
DA-MPC               & $39.2 \pm 1.9$ & $47.0 \pm 1.3$  & $44.2 \pm 0.9$ & $61.3 \pm 1.3$ \\
\midrule
BRACE-nominal-only   & $29.9 \pm 10.6$ & $34.4 \pm 11.8$ & $7.6 \pm 7.5$ & $6.4 \pm 8.5$ \\
No-belief            & $44.2 \pm 2.7$ & $60.2 \pm 4.5$  & $50.7 \pm 2.4$ & $77.2 \pm 7.6$ \\
No-residual         & $59.6 \pm 1.9$ & $79.0 \pm 2.7$  & $61.6 \pm 3.1$ & $89.3 \pm 1.2$ \\
No-reweight          & $49.4 \pm 5.4$ & $54.4 \pm 10.0$ & $53.8 \pm 5.5$ & $83.6 \pm 9.7$ \\
Frozen-nominal       & $56.0 \pm 1.6$ & $68.9 \pm 2.6$  & $60.6 \pm 2.5$ & $\mathbf{93.6 \pm 3.2}$ \\
\textbf{BRACE}       & $\mathbf{64.1 \pm 1.0}$ & $\mathbf{82.4 \pm 2.4}$
                     & $\mathbf{65.0 \pm 1.7}$ & $\mathbf{93.2 \pm 7.0}$ \\
\bottomrule
\end{tabular}}}
\ifshowrevision\par\smallskip\fi
\begingroup
\ifshowrevision\color{blue}\fi
\scriptsize
\setlength{\tabcolsep}{0pt}
\providecommand{\lb}[1]{{\color{gray}#1}}
% Option C: one policy per method on the same 2,500 fields; generated from
% BRACE/repro/optionC_all_batches_20260930_165251/tables_optionC_final.json
\begin{tabular*}{\columnwidth}{@{\extracolsep{\fill}}lcccccccc@{}}
\toprule
 & \multicolumn{4}{c}{\texttt{Quadrant}} & \multicolumn{4}{c}{\texttt{Corridor}} \\
\cmidrule(lr){2-5}\cmidrule(l){6-9}
 & \multicolumn{2}{c}{$[0.1,0.5]$} & \multicolumn{2}{c}{$[0.2,0.4]$} & \multicolumn{2}{c}{$[0.1,0.5]$} & \multicolumn{2}{c}{$[0.2,0.4]$} \\
Method & SR & \lb{$L$} & SR & \lb{$L$} & SR & \lb{$L$} & SR & \lb{$L$} \\
\midrule
Centerline & $8.6_{\pm1.2}$ & \lb{7.7} & $6.3_{\pm1.0}$ & \lb{5.5} & $0.0_{\pm0.0}$ & \lb{0.0} & $0.0_{\pm0.0}$ & \lb{0.0} \\
DA-MPC & $41.8_{\pm1.9}$ & \lb{40.2} & $48.4_{\pm2.4}$ & \lb{46.7} & $47.4_{\pm1.5}$ & \lb{45.7} & $66.8_{\pm0.8}$ & \lb{65.2} \\
RP-MPC & $45.6_{\pm2.7}$ & \lb{43.9} & $66.2_{\pm1.5}$ & \lb{64.6} & $47.2_{\pm1.9}$ & \lb{45.5} & $77.9_{\pm3.3}$ & \lb{76.5} \\
\midrule
BRACE-nominal-only & $15.9_{\pm0.8}$ & \lb{14.7} & $17.1_{\pm1.9}$ & \lb{15.9} & $0.0_{\pm0.0}$ & \lb{0.0} & $0.0_{\pm0.0}$ & \lb{0.0} \\
No-belief & $45.1_{\pm3.7}$ & \lb{43.5} & $61.8_{\pm2.8}$ & \lb{60.1} & $51.2_{\pm2.4}$ & \lb{49.5} & $87.0_{\pm2.3}$ & \lb{85.9} \\
No-residual & $57.1_{\pm2.3}$ & \lb{55.4} & $79.8_{\pm1.3}$ & \lb{78.5} & $61.1_{\pm2.7}$ & \lb{59.5} & $88.2_{\pm2.1}$ & \lb{87.1} \\
No-reweight & $53.6_{\pm2.7}$ & \lb{51.9} & $67.4_{\pm1.9}$ & \lb{65.8} & $59.8_{\pm2.0}$ & \lb{58.1} & $95.8_{\pm0.8}$ & \lb{95.0} \\
Frozen-nominal & $55.2_{\pm3.0}$ & \lb{53.5} & $69.9_{\pm1.6}$ & \lb{68.3} & $63.7_{\pm2.5}$ & \lb{62.1} & $90.9_{\pm2.5}$ & \lb{89.9} \\
\midrule
\textbf{BRACE} & $\mathbf{59.4}_{\pm2.9}$ & \lb{57.8} & $\mathbf{81.1}_{\pm2.2}$ & \lb{79.8} & $\mathbf{65.6}_{\pm2.1}$ & \lb{64.0} & $\mathbf{100.0}_{\pm0.1}$ & \lb{99.8} \\
\bottomrule
\end{tabular*}
\endgroup
\end{table}

\begin{table}[t]
\centering
\caption{\revadd{Range of $L$ (\%) over the five trained seeds.}}
\label{tab:seed_lb}
\begingroup
\ifshowrevision\color{blue}\fi
\scriptsize
\setlength{\tabcolsep}{0pt}
\begin{tabular*}{\columnwidth}{@{\extracolsep{\fill}}lcccc@{}}
\toprule
 & \multicolumn{2}{c}{\texttt{Quadrant}} & \multicolumn{2}{c}{\texttt{Corridor}} \\
\cmidrule(lr){2-3}\cmidrule(l){4-5}
Method & $[0.1,0.5]$ & $[0.2,0.4]$ & $[0.1,0.5]$ & $[0.2,0.4]$ \\
\midrule
BRACE-nominal-only & 14.7--43.2 & 15.9--52.0 & 0.0--16.7 & 0.0--20.1 \\
No-belief & 35.7--44.6 & 49.8--60.1 & 43.9--49.5 & 62.7--85.9 \\
No-residual & 48.5--56.3 & 70.5--78.7 & 50.8--60.3 & 83.2--88.3 \\
No-reweight & 41.8--53.8 & 42.1--65.8 & 43.9--58.1 & 67.5--95.0 \\
Frozen-nominal & 50.1--53.8 & 61.7--68.9 & 53.5--62.1 & 84.6--94.2 \\
\midrule
\textbf{BRACE} & 57.8--60.9 & 77.5--81.3 & 57.4--64.0 & 78.5--99.8 \\
\bottomrule
\end{tabular*}
\vspace{-0.4cm}
\endgroup
\end{table}

Table~\ref{tab:quad_main} reports success rate on two reach-avoid maps, \texttt{Quadrant} and \texttt{Corridor},
evaluated on the training wind distribution $[0.1,0.5]$ and on a narrow
\revadd{in-distribution }band $[0.2,0.4]$. \revadd{$L$ is the one-sided nominal
$95\%$ Clopper--Pearson bound of Section~\ref{sec:belief_bound} ($\beta=0.05$,
$N=2{,}500$). BRACE attains the highest success rate and $L$ in all four
conditions. Across seeds (Table~\ref{tab:seed_lb}), its $L$ varies by at most
$6.6$\,pp except in narrow-band \texttt{Corridor} ($78.5$--$99.8\%$).}

\subsubsection{\revdel{Planning vs.\ tracking compensation}
\revadd{Comparison with tracking and replanning baselines}}
\begin{figure}[t]
    \centering
    \includegraphics[width=\linewidth]{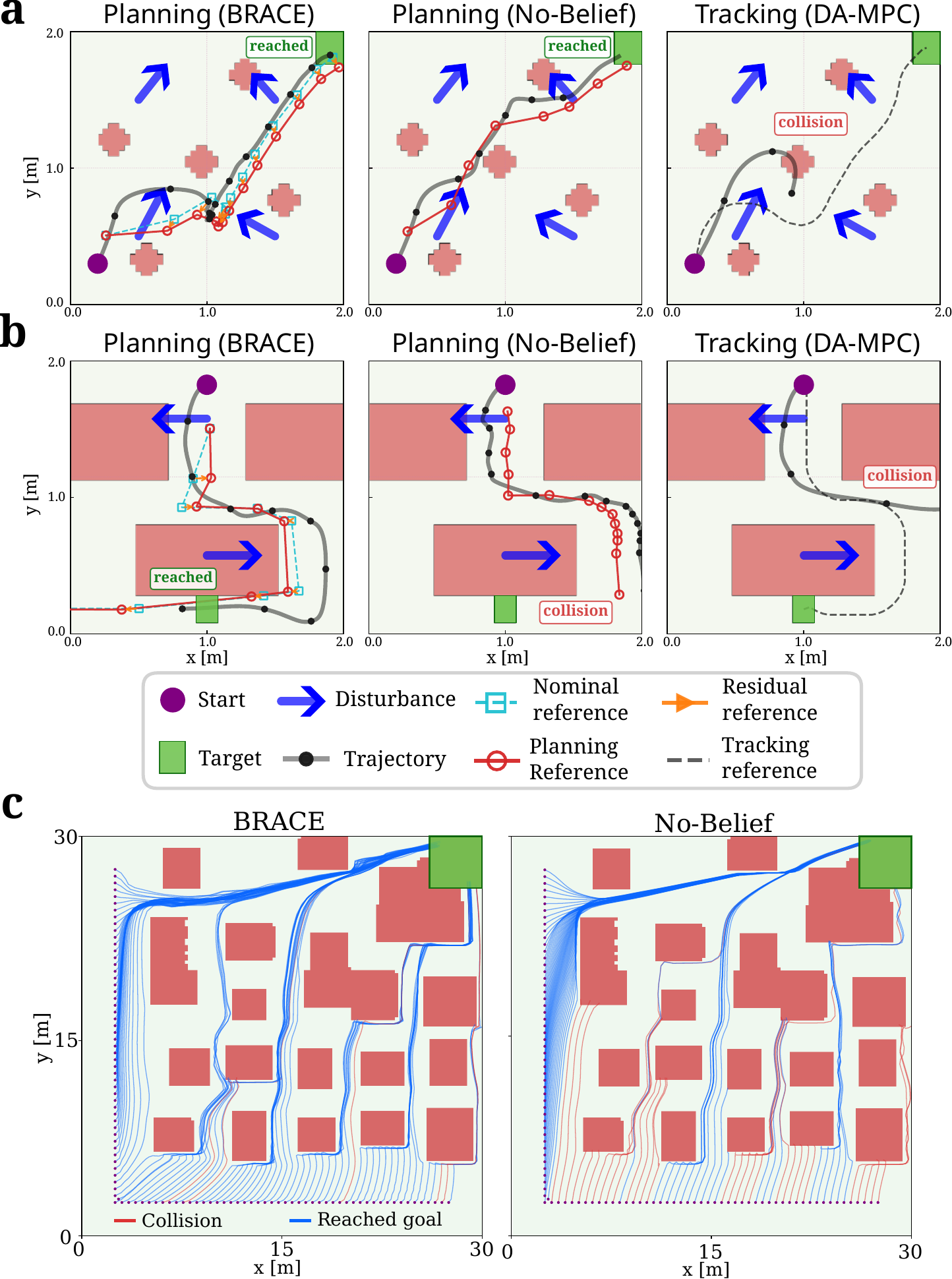}
    \vspace{-0.6cm}
    \caption{Representative trajectories.
    (a) \revdel{BRACE}\revadd{No-belief} adapts the reference geometry around the obstacle; disturbance-aware
MPC (DA-MPC) collides while tracking \revdel{an infeasible}\revadd{the fixed} reference.
    (b) BRACE selects a safer path; No-belief
    collides.
    (c) Urban evaluation rollouts for one seed: BRACE reaches 95/100 goals,
compared with 71/100 for No-belief.}
\vspace{-0.5cm}
    \label{fig:trajectory_analysis}
\end{figure}

DA-MPC receives the same IMM belief as BRACE but underperforms even
No-belief on both maps (Table~\ref{tab:quad_main}), despite the latter
having no access to belief at all. Tracking-layer compensation is
bottlenecked when the given medial-axis reference path becomes infeasible under
wind. \Cref{fig:trajectory_analysis}a illustrates this: the learned
planner changes the reference geometry and detours around the obstacle,
whereas DA-MPC stays on the prescribed reference and collides.
\revadd{RP-MPC adapts velocity references online using the IMM
disturbance mean and the same LQR tracker.
BRACE outperforms this replanning baseline by
13.9--22.0\,pp across all four conditions
(Table~\ref{tab:quad_main}).}

% \begin{figure}[t]
%     \centering
%     \includegraphics[width=\linewidth]{figure/Figure_urban.pdf}
%     \vspace{-0.7cm}
%     \caption{
% Urban evaluation rollouts for one seed: BRACE reaches 95/100 goals,
% compared with 71/100 for No-belief.
% }
%     \label{fig:urban_results}
%     \vspace{-0.6cm}
% \end{figure}

\subsubsection{Effect of belief conditioning}
Conditioning the planner on belief lifts success rate by roughly
\revdel{14--22\,pp}\revadd{13--19\,pp} over No-belief on both maps (Table~\ref{tab:quad_main}).
Belief is most useful when it changes the planned reference rather than
the tracking response. \Cref{fig:trajectory_analysis}b shows a
representative case: BRACE biases the reference toward a safer corridor
under the inferred disturbance, while No-belief selects a lower-margin
trajectory and collides.

\subsubsection{Structured residual vs.\ concatenated belief}
BRACE and No-residual use the same belief input but fuse it
differently. No-residual treats the belief as an additional
observation, whereas BRACE decomposes the mean into a nominal and disturbance-based residual term\revdel{as in}. BRACE outperforms No-residual by \revdel{3--5\,pp}\revadd{1.3--11.8\,pp}
across all four conditions (Table~\ref{tab:quad_main}). The gap is
\revdel{modest but consistent}\revadd{consistent and largest in narrow-band \texttt{Corridor}}, suggesting that the residual structure uses
belief more effectively than simple feature concatenation.
\revadd{No-belief, No-residual, and Frozen-nominal use 128-unit critic hidden layers versus 256-unit
nominal critic hidden layers in BRACE; matching this width while holding
the other hyperparameters fixed did not improve their mean success.}

\subsubsection{Credit assignment and joint training}
\begin{figure}[t]
    \centering
    \includegraphics[width=1.0\linewidth]
    {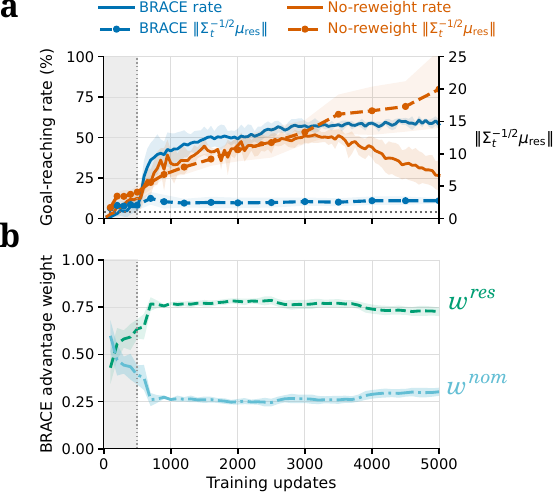}
    % \vspace{-0.2cm}
\caption{\revadd{
Advantage reweighting in \texttt{Quadrant}. (a) Goal-reaching rate (solid, left axis; five seeds) and residual mean $\|\Sigma_t^{-1/2}\mu_\mathrm{res}\|$ averaged over 10 fixed states (dashed, right axis).
(b) BRACE residual and nominal weights $w^\mathrm{res}$, $w^\mathrm{nom}$ (five seeds). Mean $\pm$ SD; gray: warmup.
}}
\vspace{-0.3cm}
    \label{fig:reweight_analysis}
\end{figure}

\revdel{Removing the likelihood-based credit assignment causes
substantial drops with much larger seed variance
(Table~\ref{tab:quad_main}), and Frozen-nominal, which trains the
residual on top of a frozen No-belief nominal, also underperforms
BRACE on three of the four conditions.}\revadd{BRACE outperforms No-reweight and Frozen-nominal in all four
conditions (Table~\ref{tab:quad_main}), and its lower bounds vary less across
seeds than No-reweight's (Table~\ref{tab:seed_lb}). The training curves in \texttt{Quadrant} suggest improved stability
(Fig.~\ref{fig:reweight_analysis}a): BRACE's goal-reaching rate
stabilizes near $60\%$, whereas No-reweight peaks near $50\%$ and then
declines while its standardized residual mean keeps growing. BRACE's
residual and nominal weights settle near $0.75$ and $0.25$--$0.30$
after warmup (Fig.~\ref{fig:reweight_analysis}b). These results indicate that likelihood-ratio advantage reweighting stabilizes joint training of the nominal and residual branches.}
\begin{figure}[t]
    \centering
    \includegraphics[width=0.8\linewidth]{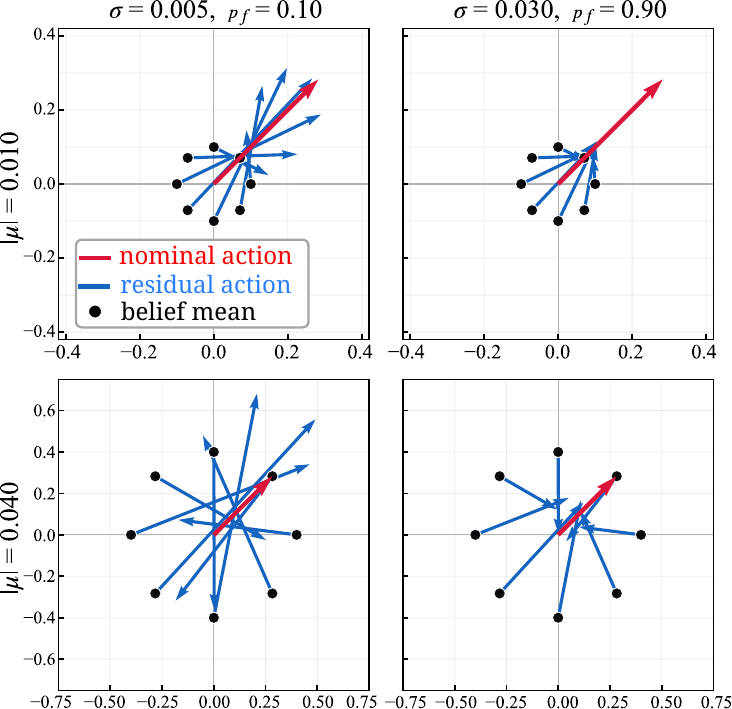}
    \vspace{-0.2cm}
    \caption{Learned residual policy in \texttt{Quadrant}, probed at a
    fixed state and nominal action. Rows vary the belief-mean magnitude
    $|(\mu_x,\mu_y)|$; columns vary the estimator \revdel{uncertainty}\revadd{uncertainty and mode probability}
    $(\sigma_x,\sigma_y,p_\mathrm{fast})$. Each subplot sweeps eight
    belief directions: black points show scaled belief means, red
    arrows the nominal action, and blue arrows the residual action.}
    \vspace{-0.4cm}
    \label{fig:residual}
\end{figure}

\subsubsection{Residual contribution and behavior}
Zeroing the residual at inference (BRACE-nominal-only) collapses
success rate to \revdel{29.9\% on \texttt{Quadrant} and 7.6\% on
\texttt{Corridor} (from 64.1\%/65.0\%).}\revadd{15.9\% on \texttt{Quadrant} and 0.0\% on
\texttt{Corridor} (from 59.4\%/65.6\%).} The residual is therefore not a
small correction on top of an independently strong nominal policy; the
two branches co-adapt into a coupled navigation strategy.

\Cref{fig:residual} probes the behavior of the learned residual policy in the \texttt{Quadrant} environment. When the estimated
disturbance is small (top row), the residual reinforces the nominal
action toward the goal. As disturbance magnitude grows (bottom row),
the residual aligns against $(\mu_x,\mu_y)$, consistent with active
wind compensation. Across columns, residual magnitude decreases \revdel{as
uncertainty grows, indicating that the policy modulates corrective
authority by estimator confidence rather than the mean disturbance
alone.}\revadd{across the uncertainty and mode-probability settings, indicating modulation beyond the disturbance mean alone.}

\subsection{Urban Wind-Field Results}
On \texttt{Urban}, BRACE reaches
$92.8 \pm 3.1$\% versus $72.6 \pm 4.7$\% for No-belief (mean $\pm$ std
over 5 seeds), which corresponds to a $20.2$\,pp gain \revadd{(\Cref{fig:trajectory_analysis}c)}. \revdel{Per-seed reach rates are
$0.90, 0.92, 0.90, 0.95, 0.97$ for BRACE and
$0.75, 0.65, 0.76, 0.71, 0.76$ for No-belief.} The worst BRACE seed
exceeds the best No-belief seed, showing \revdel{that the synthetic-wind
advantage transfers to a realistic spatial wind field.}\revadd{an advantage in this separately trained realistic spatial wind-field setting.}

\section{Hardware Experiments}

\label{sec:hardware}

\subsection{Setup}
We deploy BRACE on a Crazyflie~2.1 with motion-capture pose feedback.
Matching the simulation decomposition, the high-level policy plans at
2\,Hz and the low-level LQR tracker runs at 40\,Hz. Wind is generated
by four WORX~Turbine~600 blowers grouped into two sources (two blowers
each), forming two wind zones which are unknown to
the controller. We compute the lateral force of the applied wind offline from the flight
logs by inverse dynamics, subtracting the attitude-reconstructed thrust from the
measured horizontal acceleration; this yields a steady $0.05$--$0.15$\,N, corresponding to approximately
$0.2$--$0.6$ times the vehicle weight.

We evaluate two configurations. In \textbf{left-right} the two zones
push in opposite directions\revdel{ ($+x$, $-x$)}, requiring online adaptation
as the drone crosses the boundary; in \textbf{left-left} both push from
the same side\revdel{ ($-x$)}, requiring a persistent correction. For each, we
\revdel{run 10 BRACE trials and 10 No-belief trials (40 total).}\revadd{analyze 20 BRACE trials and 20 No-belief trials (80 total). Obstacles are virtual, and success is evaluated offline from logged positions: a trial succeeds if the drone enters the $0.15$\,m goal disk while staying outside all obstacles and within the workspace.}
\revdel{A trial succeeds if the drone reaches the goal without contacting obstacles or workspace boundaries.}
\subsection{Results}
\begin{figure}[!htbp]
    \centering
    \ifshowrevision\par\smallskip\fi
    \revadd{\includegraphics[width=\linewidth]{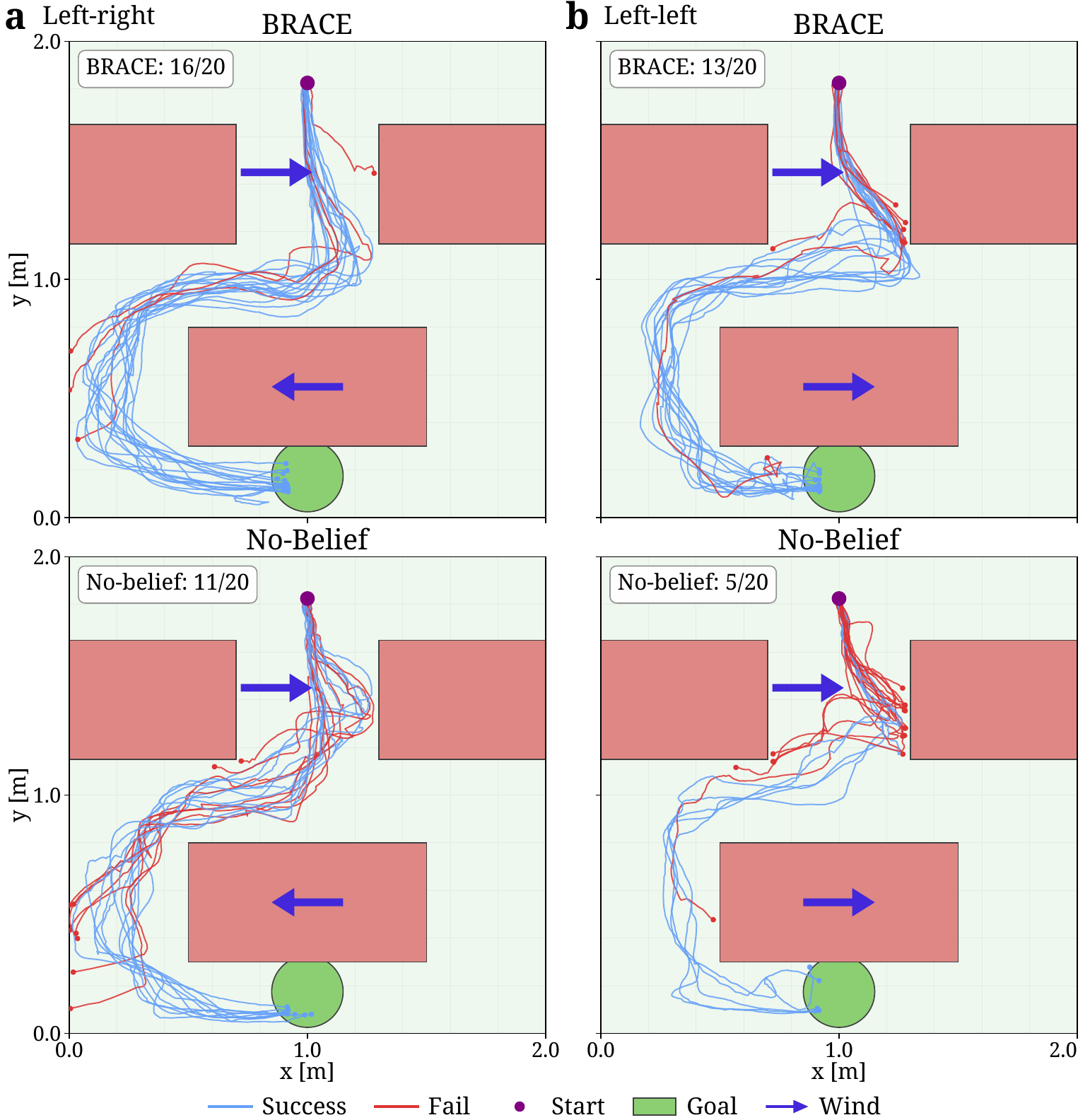}}
    % \vspace{-0.5cm}
    \caption{Hardware trajectories under two wind configurations.
    (a) Left-right configuration with opposing wind directions.
    (b) Left-left configuration with same-side wind. 
    Top row: BRACE. Bottom row: no-belief ablation. Successful trials
    are shown in blue and failures in red.}
    % \vspace{-0.5cm}
    \label{fig:hw_traj}
\end{figure}
BRACE outperforms the
No-belief ablation in both wind configurations.
In \textbf{left-right} (opposing $\pm x$ winds), BRACE \revdel{succeeds in
8/10 trials versus 6/10}\revadd{reaches the goal in 16/20 trials
versus 11/20} without belief, a \revdel{$+20$}\revadd{$+25$}\,pp gap. In
\textbf{left-left} (same-side $+x$ wind), the gap widens to
$+40$\,pp (\revdel{$7/10$ vs.\ $3/10$}\revadd{$13/20$ vs.\ $5/20$}).
\revdel{The left-left case is especially informative: the wind pushes from the
same side throughout the trajectory, so the drone must commit to a persistent one-directional
correction. Belief supplies exactly that direction, so the gap over No-belief
is widest here ($+40$\,pp).}
\revadd{The corresponding nominal two-sided 95\% Clopper--Pearson intervals
are $[56.3,94.3]\%$ for BRACE and $[31.5,76.9]\%$ for No-belief in left-right,
and $[40.8,84.6]\%$ and $[8.7,49.1]\%$ in left-left. Two-sided Fisher's exact
test shows significantly higher success for BRACE in left-left ($p=0.025$),
but not left-right ($p=0.176$), at the $0.05$ level.}

\revdel{The trajectory statistics support this
interpretation.} BRACE achieves larger \revadd{mean }%
minimum obstacle clearance in both configurations
(\revdel{$0.078 \pm 0.041$}\revadd{$0.082 \pm 0.035$}\,m vs.\
\revdel{$0.053 \pm 0.029$}\revadd{$0.077 \pm 0.037$}\,m in left-right;
\revdel{$0.047 \pm 0.035$}\revadd{$0.047 \pm 0.029$}\,m vs.\
\revdel{$0.041 \pm 0.029$}\revadd{$0.034 \pm 0.022$}\,m in left-left\revadd{, neither difference is significant,
Welch $p=0.69$ and $p=0.12$}). \revdel{with a 4--5\,cm larger
lateral offset from the windward slab in the
bottleneck region (Fig.~\ref{fig:hw_traj}).}\revdel{Belief changes the planned
trajectory geometry, not just tracking along a fixed reference. The hierarchical decomposition (belief-guided planning above a low-level LQR tracker) enabled zero-shot transfer with no platform-specific retuning.} \revadd{The learned reference generator was deployed on hardware without retraining or platform-specific retuning of its weights. }

%%%%%%%%%%%%%%%%%%%%%%%%%%%%%%%%%%%%%%%%%%%%%%%%%%%%%%%%%%%%%%%%%%%%%%%%%%%%%%%%
\section{Conclusion}
\label{sec:conclusion}
We presented BRACE, a hierarchical RL framework for reach-avoid
navigation under unknown, \revadd{bounded, }spatially varying disturbances. A CBF
feasibility-margin analysis motivates planning-layer adaptation\revdel{ by
showing how bounded actuation and disturbance direction induce
spatially varying safety margins}. BRACE realizes this principle by
decomposing high-level motion into a nominal policy and a
belief-guided residual policy that adapts the planned reference\revdel{ using an
IMM disturbance belief inferred from model prediction errors}. Likelihood-ratio
advantage reweighting \revdel{stabilizes}\revadd{supports} joint training of the nominal and
residual policies. Across three simulation environments\revdel{ and hardware experiments}, BRACE
improves reach-avoid performance over \revdel{tracking-layer compensation and RL baselines without disturbance belief}\revadd{the evaluated tracking- and planning-layer baselines}.
\revadd{While the learned DCBF does not establish closed-loop safety, we provide
high-confidence lower bounds on the reach-avoid success probability through a-posteriori Monte-Carlo simulations.}
\revdel{Several directions extend BRACE beyond its
current scope.}\revadd{Future work includes establishing formal continuous-time
safety guarantees under disturbances and
extending evaluation to}\revdel{The planar disturbance setting can be generalized to} 3D\revdel{ disturbance fields including vertical wind components. Hardware
validation can move from indoor fan-generated wind and virtual
obstacles to }\revadd{ wind fields and }outdoor environments
\revdel{with turbulent wind and }\revadd{with }physical obstacles.
\revdel{
Richer disturbance representations beyond the two-mode Gaussian IMM could
broaden applicability to multimodal or non-Gaussian
disturbance structures.}

% Final-column balancing is deferred until the page-limit edit is complete.
\bibliographystyle{IEEEtran}
\bibliography{root}

@article{hewing2020learning,
  title={Learning-based model predictive control: Toward safe learning in control},
  author={Hewing, Lukas and Wabersich, Kim P and Menner, Marcel and Zeilinger, Melanie N},
  journal={Annual Review of Control, Robotics, and Autonomous Systems},
  volume={3},
  number={1},
  pages={269--296},
  year={2020},
  publisher={Annual Reviews}
}

@article{berkenkamp2017safe,
  title={Safe model-based reinforcement learning with stability guarantees},
  author={Berkenkamp, Felix and Turchetta, Matteo and Schoellig, Angela and Krause, Andreas},
  journal={Advances in neural information processing systems},
  volume={30},
  year={2017}
}

@article{ostafew2016robust,
  title={Robust constrained learning-based NMPC enabling reliable mobile robot path tracking},
  author={Ostafew, Chris J and Schoellig, Angela P and Barfoot, Timothy D},
  journal={The International Journal of Robotics Research},
  volume={35},
  number={13},
  pages={1547--1563},
  year={2016},
  publisher={SAGE Publications Sage UK: London, England}
}

@inproceedings{singh2017robust,
  title={Robust online motion planning via contraction theory and convex optimization},
  author={Singh, Sumeet and Majumdar, Anirudha and Slotine, Jean-Jacques and Pavone, Marco},
  booktitle={2017 IEEE International Conference on Robotics and Automation (ICRA)},
  pages={5883--5890},
  year={2017},
  organization={IEEE}
}

@article{wu2024multi,
  title={Multi-objective reinforcement learning for autonomous drone navigation in urban areas with wind zones},
  author={Wu, Jiahao and Ye, Yang and Du, Jing},
  journal={Automation in construction},
  volume={158},
  pages={105253},
  year={2024},
  publisher={Elsevier}
}

@inproceedings{huang2023datt,
  title={DATT: Deep Adaptive Trajectory Tracking for Quadrotor Control},
  author={Huang, Kevin and Rana, Rwik and Spitzer, Alexander and Shi, Guanya and Boots, Byron},
  booktitle={Conference on Robot Learning},
  pages={326--340},
  year={2023},
  organization={PMLR}
}

@article{coursey2025rl,
  title={Combining Reinforcement Learning and Cascade PID Control for UAV Disturbance Rejection},
  author={Coursey, Austin and Quinones-Grueiro, Marcos and Alvarez, Luis and Biswas, Gautam},
  journal={Available at SSRN 5120263},
  year={2025}
}

@article{ahmed2023rl,
  title={A Reinforcement Learning Approach for Robust Supervisory Control of UAVs Under Disturbances},
  author={Ahmed, Ibrahim and Quinones-Grueiro, Marcos and Biswas, Gautam},
  journal={arXiv preprint arXiv:2305.12543},
  year={2023}
}

@inproceedings{johannink2019residual,
  title={Residual reinforcement learning for robot control},
  author={Johannink, Tobias and Bahl, Shikhar and Nair, Ashvin and Luo, Jianlan and Kumar, Avinash and Loskyll, Matthias and Ojea, Juan Aparicio and Solowjow, Eugen and Levine, Sergey},
  booktitle={2019 international conference on robotics and automation (ICRA)},
  pages={6023--6029},
  year={2019},
  organization={IEEE}
}

@article{zintgraf2020varibad,
  title={VariBAD: a very good method for Bayes-adaptive deep RL via meta-learning},
  author={Zintgraf, L and Shiarlis, K and Igl, M and Schulze, S and Gal, Y and Hofmann, K and Whiteson, S},
  journal={Proceedings of ICLR 2020},
  year={2020},
  publisher={International Conference on Learning Representations}
}

@inproceedings{rakelly2019efficient,
  title={Efficient off-policy meta-reinforcement learning via probabilistic context variables},
  author={Rakelly, Kate and Zhou, Aurick and Finn, Chelsea and Levine, Sergey and Quillen, Deirdre},
  booktitle={International conference on machine learning},
  pages={5331--5340},
  year={2019},
  organization={PMLR}
}

@article{wang2025hybrid,
  title={Improving Quadrotor Tracking Performance in Wind Disturbances Using Hybrid Reinforcement Learning Control},
  author={Wang, Mengyun and Liu, Yunzhuo and Niu, Yifeng and Yang, Yuqi},
  journal={Guidance, Navigation and Control},
  volume={5},
  number={01},
  pages={71--83},
  year={2025},
  publisher={World Scientific}
}

@article{zhang2024change,
  title={Why change your controller when you can change your planner: Drag-aware trajectory generation for quadrotor systems},
  author={Zhang, Hanli and Srikanthan, Anusha and Folk, Spencer and Kumar, Vijay and Matni, Nikolai},
  journal={arXiv preprint arXiv:2401.04960},
  year={2024}
}

@article{wang2025seal,
  title={Seal: Safety enhanced trajectory planning and control framework for quadrotor flight in complex environments},
  author={Wang, Yiming and Ma, Jianbin and Wu, Junda and Li, Huizhe and Zhou, Zhexuan and Gong, Youmin and Mei, Jie and Ma, Guangfu},
  journal={arXiv preprint arXiv:2503.03346},
  year={2025}
}

@article{wu2021external,
  title={External forces resilient safe motion planning for quadrotor},
  author={Wu, Yuwei and Ding, Ziming and Xu, Chao and Gao, Fei},
  journal={IEEE Robotics and Automation Letters},
  volume={6},
  number={4},
  pages={8506--8513},
  year={2021},
  publisher={IEEE}
}

@article{lapandic2024metampc,
  title={Meta-learning augmented mpc for disturbance-aware motion planning and control of quadrotors},
  author={Lapandi{\'c}, D{\v{z}}enan and Xie, Fengze and Verginis, Christos K and Chung, Soon-Jo and Dimarogonas, Dimos V and Wahlberg, Bo},
  journal={IEEE Control Systems Letters},
  volume={8},
  pages={3045--3050},
  year={2024},
  publisher={IEEE}
}

@inproceedings{yuan2025policy,
  title={Policy decorator: Model-agnostic online refinement for large policy model},
  author={Yuan, Xiu and Mu, Tongzhou and Tao, Stone and Fang, Yunhao and Zhang, Zhang and Su, Hao},
  booktitle={International Conference on Learning Representations},
  volume={2025},
  pages={28129--28164},
  year={2025}
}

@article{han2009adrc,
  title={From PID to active disturbance rejection control},
  author={Han, Jingqing},
  journal={IEEE transactions on Industrial Electronics},
  volume={56},
  number={3},
  pages={900--906},
  year={2009},
  publisher={IEEE}
}

@article{wu2024l1quad,
  title={L 1 Quad: L 1 Adaptive Augmentation of Geometric Control for Agile Quadrotors With Performance Guarantees},
  author={Wu, Zhuohuan and Cheng, Sheng and Zhao, Pan and Gahlawat, Aditya and Ackerman, Kasey A and Lakshmanan, Arun and Yang, Chengyu and Yu, Jiahao and Hovakimyan, Naira},
  journal={IEEE Transactions on Control Systems Technology},
  volume={33},
  number={2},
  pages={597--612},
  year={2025},
  publisher={IEEE}
}

@inproceedings{schmid2022real,
  title={A real-time GP based MPC for quadcopters with unknown disturbances},
  author={Schmid, Niklas and Gruner, Jonas and Abbas, Hossam S and Rostalski, Philipp},
  booktitle={2022 American control conference (ACC)},
  pages={2051--2056},
  year={2022},
  organization={IEEE}
}

@inproceedings{fan2020fast,
  title={Fast and guaranteed safe controller synthesis for nonlinear vehicle models},
  author={Fan, Chuchu and Miller, Kristina and Mitra, Sayan},
  booktitle={International Conference on Computer Aided Verification},
  pages={629--652},
  year={2020},
  organization={Springer}
}

@inproceedings{herbert2017fastrack,
  title={FaSTrack: A modular framework for fast and guaranteed safe motion planning},
  author={Herbert, Sylvia L and Chen, Mo and Han, SooJean and Bansal, Somil and Fisac, Jaime F and Tomlin, Claire J},
  booktitle={2017 IEEE 56th Annual Conference on Decision and Control (CDC)},
  pages={1517--1522},
  year={2017},
  organization={IEEE}
}

@article{schmid2026maximizing,
  title={Maximizing Reach-Avoid Probabilities for Linear Stochastic Systems via Control Architectures},
  author={Schmid, Niklas and Choi, Jaeyoun and So, Oswin and Fan, Chuchu},
  journal={arXiv preprint arXiv:2601.16290},
  year={2026}
}

@book{hovakimyan2010l1,
  title={L1 adaptive control theory: Guaranteed robustness with fast adaptation},
  author={Hovakimyan, Naira and Cao, Chengyu},
  year={2010},
  publisher={SIAM}
}

@inproceedings{seo2019robust,
  title={Robust trajectory planning for a multirotor against disturbance based on hamilton-jacobi reachability analysis},
  author={Seo, Hoseong and Lee, Donggun and Son, Clark Youngdong and Tomlin, Claire J and Kim, H Jin},
  booktitle={2019 IEEE/RSJ International Conference on Intelligent Robots and Systems (IROS)},
  pages={3150--3157},
  year={2019},
  organization={IEEE}
}

@article{romero2022timeoptimal,
  title={Time-optimal online replanning for agile quadrotor flight},
  author={Romero, Angel and Penicka, Robert and Scaramuzza, Davide},
  journal={IEEE Robotics and Automation Letters},
  volume={7},
  number={3},
  pages={7730--7737},
  year={2022},
  publisher={IEEE}
}

@article{zhou2024internal,
  title={Internal and external disturbances aware motion planning and control for quadrotors},
  author={Zhou, Xiaobin and Wang, Miao and Cui, Can and Wang, Yongchao and Xu, Chao and Gao, Fei},
  journal={IET Cyber-Systems and Robotics},
  volume={6},
  number={3},
  pages={e12122},
  year={2024},
  publisher={Wiley Online Library}
}

@inproceedings{otte2016anytime,
  title={Any-time path-planning: Time-varying wind field+ moving obstacles},
  author={Otte, Michael and Silva, William and Frew, Eric},
  booktitle={2016 IEEE international conference on robotics and automation (ICRA)},
  pages={2575--2582},
  year={2016},
  organization={IEEE}
}

@inproceedings{zhang2025discrete,
  title={Discrete GCBF proximal policy optimization for multi-agent safe optimal control},
  author={Zhang, Songyuan and So, Oswin and Black, Mitchell and Fan, Chuchu},
  booktitle={International Conference on Learning Representations},
  volume={2025},
  pages={70839--70869},
  year={2025}
}

@article{langson2004robust,
  title={Robust model predictive control using tubes},
  author={Langson, Wilbur and Chryssochoos, Ioannis and Rakovi{\'c}, SV and Mayne, David Q},
  journal={Automatica},
  volume={40},
  number={1},
  pages={125--133},
  year={2004},
  publisher={Elsevier}
}

@article{buch2021robust,
  title={Robust control barrier functions with sector-bounded uncertainties},
  author={Buch, Jyot and Liao, Shih-Chi and Seiler, Peter},
  journal={IEEE Control Systems Letters},
  volume={6},
  pages={1994--1999},
  year={2021},
  publisher={IEEE}
}

@article{iyengar2005robust,
  title={Robust dynamic programming},
  author={Iyengar, Garud N},
  journal={Mathematics of Operations Research},
  volume={30},
  number={2},
  pages={257--280},
  year={2005},
  publisher={INFORMS}
}

@article{schmid2025computing,
  title={Computing optimal joint chance constrained control policies},
  author={Schmid, Niklas and Fochesato, Marta and Li, Sarah HQ and Sutter, Tobias and Lygeros, John},
  journal={IEEE Transactions on Automatic Control},
  year={2025},
  publisher={IEEE}
}

@article{cannon2009probabilistic,
  title={Probabilistic constrained MPC for multiplicative and additive stochastic uncertainty},
  author={Cannon, Mark and Kouvaritakis, Basil and Wu, Xingjian},
  journal={IEEE Transactions on Automatic Control},
  volume={54},
  number={7},
  pages={1626--1632},
  year={2009},
  publisher={IEEE}
}

@article{chen2024probabilistic,
  title={Probabilistic constraint for safety-critical reinforcement learning},
  author={Chen, Weiqin and Subramanian, Dharmashankar and Paternain, Santiago},
  journal={IEEE Transactions on Automatic Control},
  volume={69},
  number={10},
  pages={6789--6804},
  year={2024},
  publisher={IEEE}
}

@article{cosner2023robust,
  title={Robust safety under stochastic uncertainty with discrete-time control barrier functions},
  author={Cosner, Ryan K and Culbertson, Preston and Taylor, Andrew J and Ames, Aaron D},
  journal={arXiv preprint arXiv:2302.07469},
  year={2023}
}

@article{hewing2019cautious,
  title={Cautious model predictive control using gaussian process regression},
  author={Hewing, Lukas and Kabzan, Juraj and Zeilinger, Melanie N},
  journal={IEEE Transactions on Control Systems Technology},
  volume={28},
  number={6},
  pages={2736--2743},
  year={2019},
  publisher={IEEE}
}

@article{kogel2025safe,
  title={Safe hierarchical model predictive control and planning for autonomous systems},
  author={K{\"o}gel, Markus and Ibrahim, Mohamed and Kallies, Christian and Findeisen, Rolf},
  journal={International Journal of Robust and Nonlinear Control},
  volume={35},
  number={7},
  pages={2658--2676},
  year={2025},
  publisher={Wiley Online Library}
}

@article{stamouli2026layered,
  title={Layered Control of Partially Observed Stochastic Systems},
  author={Stamouli, Charis and Tsiamis, Anastasios and Pappas, George J},
  journal={arXiv preprint arXiv:2604.11956},
  year={2026}
}

@inproceedings{kolmanovsky2014reference,
  title={Reference and command governors: A tutorial on their theory and automotive applications},
  author={Kolmanovsky, Ilya and Garone, Emanuele and Di Cairano, Stefano},
  booktitle={2014 American Control Conference},
  pages={226--241},
  year={2014},
  organization={IEEE}
}

@article{blom1988interacting,
  title={The interacting multiple model algorithm for systems with Markovian switching coefficients},
  author={Blom, Henk AP and Bar-Shalom, Yaakov},
  journal={IEEE transactions on Automatic Control},
  volume={33},
  number={8},
  pages={780--783},
  year={1988},
  publisher={IEEE}
}

@article{so2024solving,
  title={Solving minimum-cost reach avoid using reinforcement learning},
  author={So, Oswin and Ge, Cheng and Fan, Chuchu},
  journal={Advances in Neural Information Processing Systems},
  volume={37},
  pages={30951--30984},
  year={2024}
}

@book{agresti2013categorical,
  title={Categorical data analysis},
  author={Agresti, Alan},
  year={2013},
  publisher={John Wiley \& Sons}
}

@article{zuliani2025closed,
  title={Closed-loop performance optimization of model predictive control with robustness guarantees},
  author={Zuliani, Riccardo and Balta, Efe C and Lygeros, John},
  journal={European Journal of Control},
  pages={101319},
  year={2025},
  publisher={Elsevier}
}

@inproceedings{gao2025learning,
  title={Learning-based Homothetic Tube MPC},
  author={Gao, Yulong and Yan, Shuhao and Zhou, Jian and Cannon, Mark},
  booktitle={2025 European Control Conference (ECC)},
  pages={2038--2043},
  year={2025},
  organization={IEEE}
}

@inproceedings{abdul2025nonlinear,
  title={Nonlinear robust optimization for planning and control},
  author={Abdul, Arshiya Taj and Saravanos, Augustinos D and Theodorou, Evangelos A},
  booktitle={2025 IEEE 64th Conference on Decision and Control (CDC)},
  pages={3383--3390},
  year={2025},
  organization={IEEE}
}

@INPROCEEDINGS{oshin2024robust, 
    AUTHOR    = {Alex Oshin AND Hassan Almubarak AND Evangelos Theodorou}, 
    TITLE     = {{Differentiable Robust Model Predictive Control}}, 
    BOOKTITLE = {Proceedings of Robotics: Science and Systems}, 
    YEAR      = {2024}, 
    ADDRESS   = {Delft, Netherlands}, 
    MONTH     = {July}, 
    DOI       = {10.15607/RSS.2024.XX.003} 
}

@article{ames2016control,
  title={Control barrier function based quadratic programs for safety critical systems},
  author={Ames, Aaron D and Xu, Xiangru and Grizzle, Jessy W and Tabuada, Paulo},
  journal={IEEE Transactions on Automatic Control},
  volume={62},
  number={8},
  pages={3861--3876},
  year={2016},
  publisher={IEEE}
}

@article{choi2026safe,
  title={Safe and Scalable Multi-Drone Payload Transport Via CBF-Based Reinforcement Learning With Zero-Shot Sim-to-Real Transfer},
  author={Choi, Jaeyoun and So, Oswin and Zhang, Songyuan and Taylor, Cooper and Fan, Chuchu},
  journal={IEEE Robotics and Automation Letters},
  year={2026},
  publisher={IEEE}
}

@book{lavalle2006planning,
  title={Planning algorithms},
  author={LaValle, Steven M},
  year={2006},
  publisher={Cambridge university press}
}

@INPROCEEDINGS{saravanos2024distributed,
  author={Saravanos, Augustinos D. and Balci, Isin M. and Bakolas, Efstathios and Theodorou, Evangelos A.},
  booktitle={2024 IEEE/RSJ International Conference on Intelligent Robots and Systems (IROS)}, 
  title={Distributed Model Predictive Covariance Steering}, 
  year={2024},
  volume={},
  number={},
  pages={5740-5747},
  doi={10.1109/IROS58592.2024.10802104}}

\end{document}